\documentclass{article} 

\usepackage{tgpagella} 
\usepackage{mathpazo}  

\usepackage[T1]{fontenc}    
\usepackage[utf8]{inputenc} 
\usepackage{microtype}      
\usepackage{inconsolata}    
\usepackage{graphicx}       
\usepackage{etoolbox}       

\usepackage{tgpagella} 
\usepackage{mathpazo}  

\usepackage{fancyhdr}
\usepackage{natbib}

\setcitestyle{authoryear,round,citesep={;},aysep={,},yysep={;}}

\newif\ifcolmfinal
\colmfinalfalse
\def\colmfinalcopy{\colmfinaltrue}

\def\addcontentsline#1#2#3{}

\makeatletter
\def\maketitle{\par
\begingroup
   \def\thefootnote{\fnsymbol{footnote}}
   \def\@makefnmark{\hbox to 0pt{$^{\@thefnmark}$\hss}} 
   \long\def\@makefntext##1{\parindent 1em\noindent
                            \hbox to1.8em{\hss $\m@th ^{\@thefnmark}$}##1}
   \@maketitle \@thanks
\endgroup
\setcounter{footnote}{0}
\let\maketitle\relax \let\@maketitle\relax
\gdef\@thanks{}\gdef\@author{}\gdef\@title{}\let\thanks\relax}

\usepackage{graphicx}
\usepackage{fancyhdr}
\def\@maketitle{\vbox{\hsize\textwidth
{\Large\bf \@title\par}
\ifcolmfinal
    \globaldefs=1\relax 
    \lhead{\raisebox{0.1cm}[0pt][0pt]{Technical Report}} 
        \rhead{\raisebox{0.1cm}[0pt][0pt]{Alibaba Token Hub, Alibaba Group}}
        \globaldefs=0\relax
    \def\And{\end{tabular}\hfil\linebreak[0]\hfil
            \begin{tabular}[t]{l}\bf\rule{\z@}{24pt}\ignorespaces}%
  \def\AND{\end{tabular}\hfil\linebreak[4]\hfil
            \begin{tabular}[t]{l}\bf\rule{\z@}{24pt}\ignorespaces}%
    \begin{tabular}[t]{l}\bf\rule{\z@}{24pt}\@author\end{tabular}%
\else
       \lhead{Under review as a conference paper at COLM 2024}
   \def\And{\end{tabular}\hfil\linebreak[0]\hfil
            \begin{tabular}[t]{l}\bf\rule{\z@}{24pt}\ignorespaces}%
  \def\AND{\end{tabular}\hfil\linebreak[4]\hfil
            \begin{tabular}[t]{l}\bf\rule{\z@}{24pt}\ignorespaces}%
    \begin{tabular}[t]{l}\bf\rule{\z@}{24pt}Anonymous authors\\Paper under double-blind review\end{tabular}%
\fi
\vskip 0.3in minus 0.1in}}

\renewenvironment{abstract}{\vskip.075in\centerline{\large\bf
Abstract}\vspace{-0.5ex}\begin{center}\begin{minipage}{0.88\textwidth}}{\end{minipage}\end{center}\par\vskip 1ex}

\def\section{\@startsection {section}{1}{\z@}{-2.0ex plus
    -0.5ex minus -.2ex}{1.5ex plus 0.3ex
minus0.2ex}{\large\bf\raggedright}}

\def\subsection{\@startsection{subsection}{2}{\z@}{-1.8ex plus
-0.5ex minus -.2ex}{0.8ex plus .2ex}{\normalsize\bf\raggedright}}
\def\subsubsection{\@startsection{subsubsection}{3}{\z@}{-1.5ex
plus      -0.5ex minus -.2ex}{0.5ex plus
.2ex}{\normalsize\raggedright}}
\def\paragraph{\@startsection{paragraph}{4}{\z@}{1.5ex plus
0.5ex minus .2ex}{-1em}{\normalsize\bf}}
\def\subparagraph{\@startsection{subparagraph}{5}{\z@}{1.5ex plus
  0.5ex minus .2ex}{-1em}{\normalsize}}

\skip\footins 9pt plus 4pt minus 2pt
\def\footnoterule{\kern-3pt \hrule width 12pc \kern 2.6pt }
\leftmargini\leftmargin \leftmarginii 2em
\def\@listi{\leftmargin\leftmargini}
\def\@listii{\leftmargin\leftmarginii
   \labelwidth\leftmarginii\advance\labelwidth-\labelsep
   \topsep 2pt plus 1pt minus 0.5pt
   \parsep 1pt plus 0.5pt minus 0.5pt
   \itemsep \parsep}
\def\@listiii{\leftmargin\leftmarginiii
    \labelwidth\leftmarginiii\advance\labelwidth-\labelsep
    \topsep 1pt plus 0.5pt minus 0.5pt
    \parsep \z@ \partopsep 0.5pt plus 0pt minus 0.5pt
    \itemsep \topsep}
\def\@listiv{\leftmargin\leftmarginiv
     \labelwidth\leftmarginiv\advance\labelwidth-\labelsep}
\def\@listv{\leftmargin\leftmarginv
     \labelwidth\leftmarginv\advance\labelwidth-\labelsep}
\def\@listvi{\leftmargin\leftmarginvi
     \labelwidth\leftmarginvi\advance\labelwidth-\labelsep}

\belowdisplayskip \abovedisplayskip
\def\normalsize{\@setsize\normalsize{11pt}\xpt\@xpt}
\def\small{\@setsize\small{10pt}\ixpt\@ixpt}
\def\footnotesize{\@setsize\footnotesize{10pt}\ixpt\@ixpt}
\def\scriptsize{\@setsize\scriptsize{8pt}\viipt\@viipt}
\def\tiny{\@setsize\tiny{7pt}\vipt\@vipt}
\def\large{\@setsize\large{14pt}\xiipt\@xiipt}
\def\Large{\@setsize\Large{16pt}\xivpt\@xivpt}
\def\LARGE{\@setsize\LARGE{20pt}\xviipt\@xviipt}
\def\huge{\@setsize\huge{23pt}\xxpt\@xxpt}
\def\Huge{\@setsize\Huge{28pt}\xxvpt\@xxvpt}

\makeatother
\AtBeginEnvironment{abstract}{\linespread{1}\selectfont}
\AtEndEnvironment{abstract}{\linespread{1.05}\selectfont}

\usepackage[colorlinks = true,
            linkcolor = blue,
            urlcolor  = blue,
            citecolor = blue,
            anchorcolor = blue]{hyperref}
\usepackage{url}            

\usepackage{booktabs}       
\usepackage{amsmath}
\usepackage{amssymb}
\usepackage{amsfonts}
\usepackage{amsthm}
\newtheorem{theorem}{Theorem}
\newtheorem{definition}{Definition}
\usepackage{multirow}
\usepackage{makecell}
\usepackage{array}
\usepackage{colortbl}
\usepackage{xcolor}         
\usepackage{subcaption}
\usepackage{wrapfig}
\usepackage{float}
\usepackage[shortlabels]{enumitem}
\usepackage[htt]{hyphenat}
\usepackage[linesnumbered,ruled,vlined]{algorithm2e}
\usepackage{tcolorbox}
\tcbuselibrary{breakable,skins}
\tcbset{before upper={\sloppy\hyphenpenalty=50\tolerance=9999\emergencystretch=2em}}

\definecolor{shadecolor}{RGB}{237,237,237}
\definecolor{backblue}{RGB}{210,230,250}

\title{Rollout-Time Intervention:\\ Unlocking Agentic Reinforcement Learning Beyond Policy Limits}

\author{Xi Chen$^{1}$, Ruixue Ding$^{1}$\thanks{Project leader}, Zhigaoyuan Wang$^{3}$, Fanrui Zhang$^{1}$, Qiang Zhang$^{1}$, Boli Chen$^{1}$,\\
\bf Shihang Wang$^{1}$, Tianfu Wang$^{3}$, Ailin Ren$^{2}$, Pengyong Wang$^{2}$, Pengjun Xie$^{1}$, Hui Xiong$^{3}$\\
\\
$^{1}$Alibaba Token Hub, Alibaba Group $^{2}$Amap, Alibaba Group $^{3}$HKUST (Guangzhou)
}

\colmfinalcopy

\begin{document}

\maketitle



\begin{abstract}

Reinforcement learning (RL) has emerged as a key paradigm for improving large language model (LLM) agents, enabling stronger reasoning, planning, and tool-use in interactive environments. However, existing agent RL methods remain fundamentally constrained by an on-policy reasoning bottleneck: when tasks exceed the model’s current capability boundary, effective trajectories become extremely sparse and training often collapses.
We introduce \textbf{Rollout-Time Intervention (RTI)}, a technique designed to elicit successful trajectories beyond the policy's intrinsic exploration capacity.
RTI operates by leveraging reward-side evidence (i.e., evaluation rubrics) during rollout to convert erroneous trajectories into successful ones, providing effective learning signals for tasks that the policy rarely solves on its own. 
Although intervention trajectories are inherently off-policy, we find that they can effectively guide policy optimization when trajectories with excessive distributional shifts are filtered out. 
Extensive experiments across five benchmarks demonstrate that RTI consistently outperforms RL baselines, enabling the post-trained policy to approach the performance of substantially stronger contemporary LLMs and, more importantly, transcend its original capability boundary.
The code is available at \url{https://github.com/Alibaba-NLP/qqr}.

\end{abstract}

\section{Introduction}
LLMs are rapidly evolving from static instruction-following systems into interactive agents that can collaborate with external environments to achieve goals~\citep{team2025kimi,froger2025scaling}. RL has been a key driver of this shift, enabling advanced capabilities such as self-reflection and long-horizon planning~\citep{guo2025deepseek,qian2025toolrl}. These advances have fueled strong performance in challenging agent settings, including mathematical reasoning with code interpreters~\citep{feng2025retoolreinforcementlearningstrategic}, deep research with search engines~\citep{jin2025searchr,sun2025zerosearch}, and complex real-world tool use~\citep{qian2025toolrl,gao2026self,yang2026topocurate,dong2026agent,li2026deepagent}.

\begin{figure}[htbp]
    \centering
    \includegraphics[width=\textwidth]{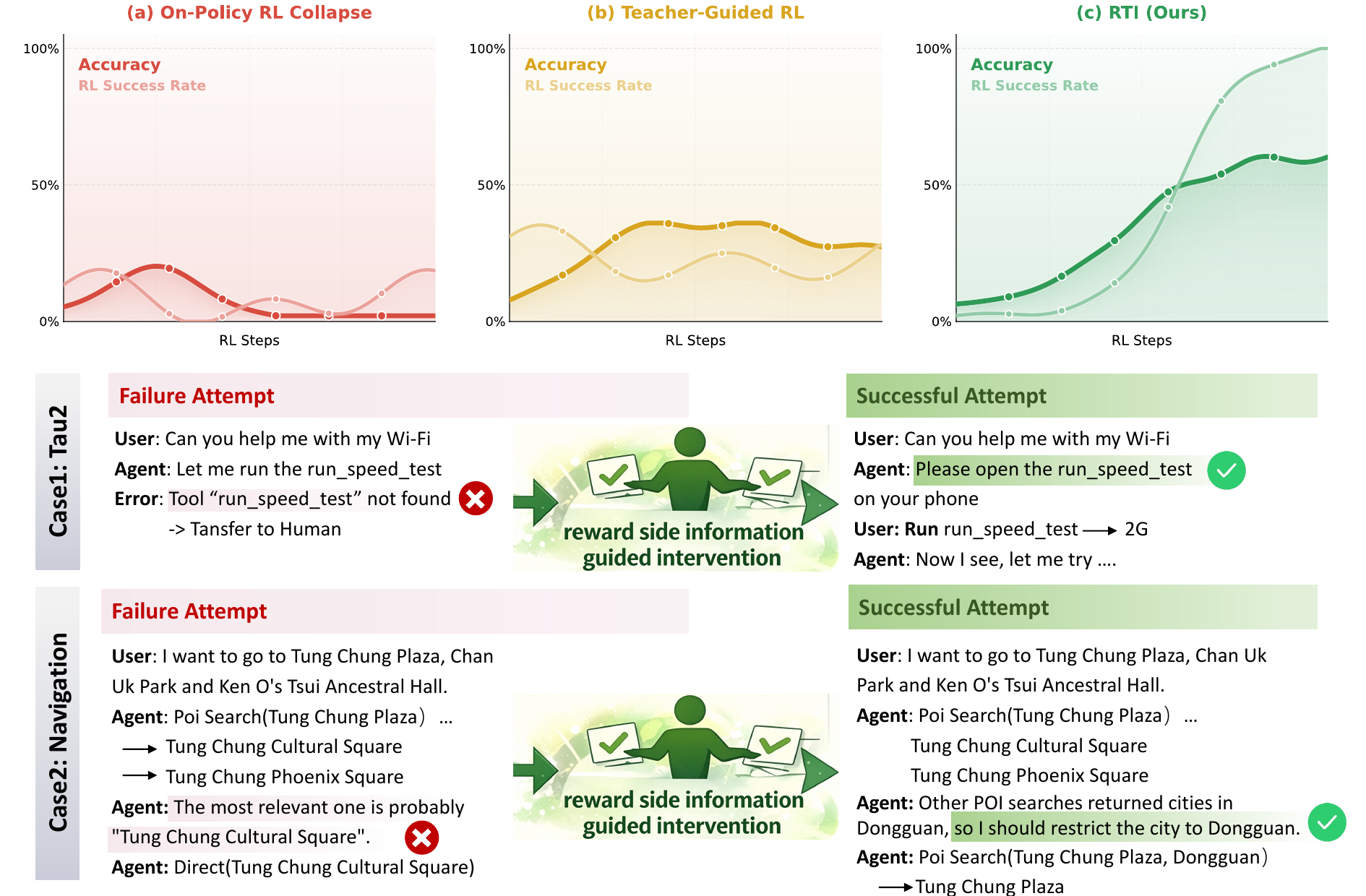}
    \caption{RTI enables successful trajectories where on-policy and teacher-guided RL fail.}
    \label{fig:intro}
\end{figure}

Nevertheless, standard RL is limited by its on-policy nature: it mainly amplifies behaviors already supported by the current policy, rather than inducing new reasoning patterns~\citep{zhao2025echo,chen2026does}. As a result, when tasks lie beyond the model’s capability boundary, successful trajectories become extremely sparse, leading to unstable training or even collapse (Figure~\ref{fig:intro}(a)). Teacher-guided methods such as LUFFY~\citep{luffy} and OPD~\citep{lu2025onpolicydistillation} can partially mitigate this issue, but their improvements remain bounded by teacher capability. 

To further understand these failures, we examine failed agent trajectories and find a recurring pattern: the agent often possesses the knowledge and tools needed to complete the task, but makes a small early reasoning or action error at a task-critical step.
As illustrated in Figure~\ref{fig:intro}, in $\tau^2$-Telecom, the agent must guide the user to invoke a user-side tool, yet the model may incorrectly treat it as an agent-accessible tool and repeatedly attempt to call it; in Navigation, user queries are often ambiguous, and the agent must interpret noisy POI search results in the context of the user's intent rather than blindly treating a returned POI as the target. In both cases, a single incorrect interaction strategy can steer subsequent reasoning into a failure loop, while correcting this early decision is often sufficient for the remaining trajectory to proceed normally.
These observations motivate us: \textit{if such early reasoning errors can be corrected so that the rollout succeeds, can RL learn from these corrected trajectories and thereby push the policy beyond its original reasoning limits (Figure~\ref{fig:intro}(c))?}

However, such intervention introduces a central technical challenge: the corrected trajectories are no longer sampled from the original policy, but generated by an intervention-induced behavior policy. This discrepancy poses two fundamental risks. First, privileged reward-side information may leak into the reasoning process, producing shortcut-driven or behaviorally implausible trajectories that could not arise under genuine task interaction. Second, intervention may induce substantial distributional shifts from the original policy. In PPO-style optimization, including GRPO, such shifts can push importance ratios beyond the clipping range, suppressing otherwise useful learning signals, while low-probability actions may yield high-variance gradients that destabilize training.

To this end, we propose \textbf{Rollout-Time Intervention (RTI)}, comprising three components: (1) \textit{reward-aware multi-agent intervention}, which employs a Validator--Corrector--Verifier pipeline to transform failed rollouts into successful intervention trajectories while preventing privileged-information leakage; (2) \textit{perplexity-based distribution gating}, which filters intervened trajectories that deviate excessively from the current policy; and (3) \textit{reasoning-shift-aware token reweighting}, which concentrates policy updates on critical tokens where intervention induces substantial strategy changes. Together, these components enable stable learning from successful trajectories beyond the policy's unaided exploration capacity, thereby pushing the policy beyond its original capability boundary.
Experiments on public and real-world benchmarks show that RTI substantially outperforms baselines and achieves a breakthrough in the boundaries of reasoning capabilities.
\section{Related Work}

\noindent \textbf{Agentic Reasoning for LLMs.} 
Agentic reasoning reframes LLMs as autonomous agents that combine planning, acting, and learning through interaction~\citep{wei2026agentic}, as exemplified by ReAct~\citep{yao2023react}, tool-use and search-based agents, and workflow-oriented systems with verifiable sub-tasks~\citep{yao2023react,jin2025searchr,sun2025zerosearch,feng2025retoolreinforcementlearningstrategic}.
This paradigm requires agents to plan over long horizons, navigate partial observability, and actively improve through feedback~\citep{wang2024survey}.
Existing methods for agent training are as follows:

\textit{(i) Data-synthesis-driven SFT} improves tool use by generating interaction trajectories for supervised fine-tuning, including ToolLLM~\citep{qin2023toolllm}, ToolACE~\citep{liu2025toolace}, APIGen-MT~\citep{prabhakar2026apigenmt}, TopoCurate~\citep{yang2026topocurate}, and simulation or self-evolving pipelines for multi-turn data generation~\citep{li2025simulating,gao2026self}. These methods are effective but depend on large-scale trajectory synthesis. 
\textit{(ii) Pure RL with verifiable rewards} has also shown promise when outcomes are easy to verify. ReTool~\citep{feng2025retoolreinforcementlearningstrategic} incorporates code execution for competition-level problem solving, Search-R1~\citep{jin2025searchr} learns search-engine use with emergent multi-query behavior, ZeroSearch~\citep{sun2025zerosearch} replaces real retrieval with LLM simulation, and ToolRL~\citep{qian2025toolrl} shows that fine-grained format and outcome rewards can support stable multi-step tool-use RL. More recently, AEPO~\citep{aepo} further improves optimization stability under long-horizon, multi-turn tool use through entropy-aware policy updates. 


\noindent \textbf{Teacher-Guided Training for LLMs.}
When tasks lie beyond the model's capability boundary, successful rollouts become extremely sparse and the policy is rarely exposed to truly corrective trajectories.
To alleviate this, recent work studies teacher-guided training in two forms. One line provides \emph{token-level credit assignment} using stronger teachers, often via trajectory-level teacher-student KL to derive dense rewards~\citep{gu2024minillm,lu2025onpolicydistillation}. Another line introduces \emph{off-policy successful trajectories}: LUFFY~\citep{luffy}, for instance, combines offline expert data with online rollouts and uses importance sampling to handle distribution shift. 

While these methods can improve exploration and learning, they all fundamentally depend on stronger teacher models, making their gains bounded by the teacher’s capability ceiling.

\section{Preliminaries}
\label{sec:preliminary}


\indent \textbf{Problem Setting.}
We formulate the agentic task as a trajectory generation problem. Let $\mathcal{T}$ denote the set of tools. Given a query $x$ sampled from a task distribution $\mathcal{D}$, the policy $\pi_\theta$ produces a multi-step interaction trajectory $\tau$ consisting of interleaved chain-of-thought reasoning $z_k$, tool invocations $a_k \in \mathcal{T}$, environmental feedback $o_k$, and a final answer $y$: $\tau = [z_1, a_1, o_1, \dots, z_k, a_k, o_k, \dots, y]$.
A verifiable reward function $r(x,\tau)\in\{0,1\}$ evaluates the trajectory.

\indent \textbf{Group Relative Policy Optimization.}
GRPO~\citep{shao2024deepseekmath} is a widely adopted RL algorithm. For each query $x$, GRPO samples a group of $G$ trajectories $\{\tau_i\}_{i=1}^G \sim \pi_\theta(\cdot \mid x)$ and normalizes the rewards into per-sample advantages:
\begin{equation}
\small
A_i = \frac{r(x,\tau_i) - \mu_r}{\sigma_r + \epsilon},\quad \mu_r = \tfrac{1}{G}\textstyle\sum_{i=1}^G r(x,\tau_i),\quad \sigma_r = \sqrt{\tfrac{1}{G}\textstyle\sum_{i=1}^G (r(x,\tau_i)\!-\!\mu_r)^2}.
\end{equation}

The policy is then updated by maximizing the clipped surrogate objective $\mathcal{J}_{\mathrm{GRPO}}(\theta)$:
\begin{equation}
\small
\label{eq:grpo}
\begin{aligned}
\mathbb{E}_{x,\{\tau_i\}}\!\Bigg[
\frac{1}{G}\sum_{i=1}^{G}\frac{1}{|\tau_i|}\sum_{t=1}^{|\tau_i|}
\min\!\Big(
\rho_{\mathrm{grpo}, i}^t A_i, \mathrm{clip}(\rho_{\mathrm{grpo}, i}^t, 1-\epsilon_{\mathrm{clip}}, 1+\epsilon_{\mathrm{clip}})\, A_i
\Big)
\Bigg]
- \beta \mathbb{D}_{\mathrm{KL}}(\pi_\theta\|\pi_\mathrm{ref}),
\end{aligned}
\end{equation}
where $\rho_{\mathrm{grpo}, i}^t = \frac{\pi_\theta(\tau_i^t \mid x, \tau_i^{<t})}{\pi_\mathrm{old}(\tau_i^t \mid x, \tau_i^{<t})}$ is the per-token probability ratio between the current policy and the behavior policy $\pi_\mathrm{old}$ that generated $\tau_i$, and $\beta > 0$ controls the strength of the KL regularization.

\section{Rollout-Time Intervention}
\label{sec:method}

\begin{figure}[t]
    \centering
    \includegraphics[width=\textwidth]{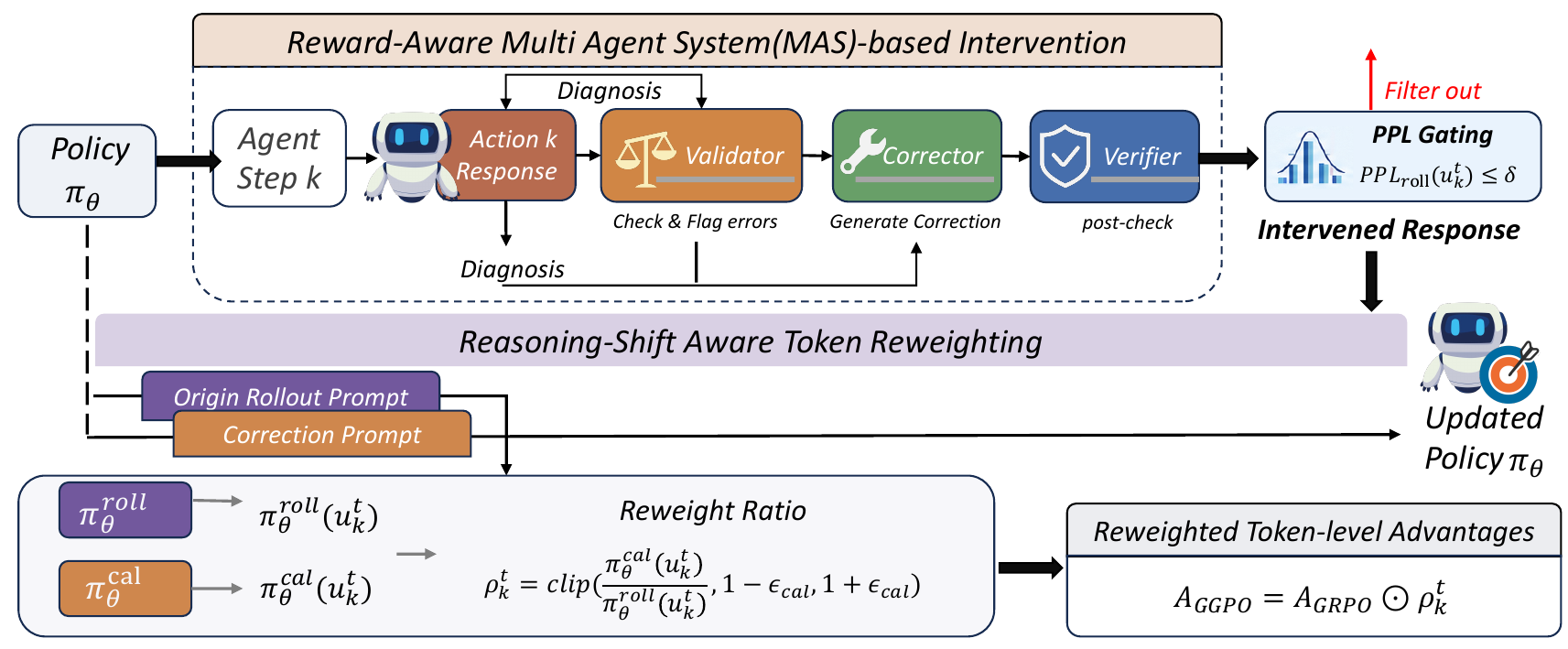}
    \vspace{-7mm}
    \caption{Overview of the proposed RTI framework.}
    \vspace{-5mm}
    \label{fig:method}
\end{figure}

To overcome the exploration bottleneck of on-policy RL, we propose \textbf{RTI}, a technique comprising three components as shown in Figure~\ref{fig:method}. First, \emph{reward-aware multi-agent system (MAS) intervention} calibrates task-critical errors at the step level, transforming failed rollouts into successful intervention trajectories while preventing privileged-information leakage. Second, \emph{perplexity-based distribution gating} filters interventions that deviate excessively from the rollout policy, mitigating ineffective or unstable updates caused by severe off-policy shifts. Third, \emph{reasoning-shift-aware token reweighting} emphasizes the critical tokens responsible for intervention-induced strategy changes, enabling more precise credit assignment. Together, these components enable stable learning from successful trajectories beyond the policy's unaided exploration capacity.

\subsection{Reward-Aware MAS-Based Intervention}
\label{sec:mas_intervention}

Given a query $x$, only when the entire group fails (i.e., all $G$ trajectories generated by $\pi_{\theta}$ receive zero reward), we invoke the intervention pipeline, an MAS consisting of a \emph{Reward-Oriented Trajectory Validator}, a \emph{Policy-Aligned Trajectory Corrector}, and an \emph{Integrity-Constrained Intervention Verifier}. By default, all three MAS roles are instantiated with the policy model itself.

\paragraph{Reward-Oriented Trajectory Validator}

The validator identifies whether the current rollout step has \emph{harmfully deviated} from a valid path to the gold outcome. Given the query $x$ and the observable history before step $k$: $h_{1:k-1} = (z_1, a_1, o_1, \dots, z_{k-1}, a_{k-1}, o_{k-1})$, the current response $\hat{u}_k$, and the reward-side evidence $\ell$, it determines whether $\hat{u}_k := (z_k, a_k)$ is still compatible with successful task completion. Importantly, the validator is \emph{reward-oriented} rather than process-prescriptive: it does not penalize suboptimal but recoverable decisions, alternative valid action orders, or stylistic differences, and instead flags only those errors that are likely to derail the trajectory. 
In this way, the validator detects only \emph{task-critical} mistakes, matching our central motivation that many rollout failures arise from early local errors rather than missing end-task capability. The prompt is shown below:

\begin{tcolorbox}[colback=gray!5,colframe=black!60,title=\textsc{Validator} prompt (schematic),fonttitle=\bfseries\small,boxrule=0.4pt,arc=1pt,left=3pt,right=3pt,top=2pt,bottom=2pt]
\scriptsize
\label{fig:prompt_judge_box}
\textbf{Inputs.} Observable history $h_{1:k-1}$; current response $\hat{u}_k$; reward-side evidence $\ell$.\\[2pt]
\textbf{Body.} \textit{(i)} \textit{Observable evidence} listing prior turns and tool outputs; \textit{(ii)} \textit{Required actions} derived from $\ell$ separated into agent-side and (for telecom) user-side groups; \textit{(iii)} \textit{Hard rules}: false-claim, ...; \textit{(iv)} \textit{Negative space} explicitly disallowing invalidation due to suboptimal-but-recoverable order.\\[2pt]
\textbf{Output.} JSON $\{\texttt{is\_valid},\,\texttt{error\_types},\,\texttt{diagnosis},\,\texttt{suggest\_next}\}$.
\end{tcolorbox}

The validator returns a structured judgment including whether the step is valid, the error type, a brief diagnosis, and a next-step suggestion. The diagnosis explains why the current step is inconsistent with the gold trajectory, while the suggested next step only indicates the direction of repair.

\paragraph{Policy-Aligned Trajectory Corrector}
When the validator detects a harmful deviation, the corrector rewrites the current step into an intervened response $\tilde{u}_k$. The key requirement is that the corrected step should remain a plausible continuation of the agent's own reasoning process. Concretely, the corrector takes as input the history $h_{1:k-1}$, tool schema $\mathcal{T}$, the original draft $\hat{u}_k$, and the validator feedback, and produces a revised response that fixes the critical error while preserving the original response style, action format, and local reasoning structure as much as possible.
\begin{tcolorbox}[colback=gray!5,colframe=black!60,title=\textsc{Corrector} prompt (schematic),fonttitle=\bfseries\small,boxrule=0.4pt,arc=1pt,left=3pt,right=3pt,top=2pt,bottom=2pt]
\scriptsize
\label{fig:prompt_corrector_box}
\textbf{Inputs.} History $h_{1:k-1}$; tool schemas $\mathcal{T}$; prior draft $\hat{u}_k$ (style reference); validator feedback (diagnosis and suggested next step); reward-side evidence $\ell$ (privileged reference, never echoed).\\[2pt]
\textbf{Body.} \textit{(i)}``silent calibrator'' specification; \textit{(ii)} four generation principles---\emph{style alignment}, \emph{progressive correction}, \emph{leakage suppression}, \emph{minimal intervention}; \textit{(iii)} domain-specific requirements; \textit{(iv)} explicit instruction that the intervened response must be derivable from visible history alone.\\[2pt]
\textbf{Output.} JSON $\{\texttt{reasoning},\,\texttt{tool\_calls},\,\texttt{content}\}$.
\end{tcolorbox}

Our goal is not to replace the trajectory with an oracle solution, but to generate a minimally corrected version that stays close to the policy's native behavior while removing the decisive error. Therefore, the corrector is instructed to follow the visible history and imitate the original reasoning pattern. 

\paragraph{Integrity-Constrained Intervention Verifier}
Finally, the verifier checks whether the intervened response is suitable. Given the original context and the corrected response $\tilde{u}_k$, it verifies three properties: (i) the correction remains consistent with the gold requirement, (ii) it does not leak privileged information, and (iii) it is still a reasonable continuation of the original policy behavior. 
\begin{tcolorbox}[colback=gray!5,colframe=black!60,title=\textsc{Verifier} prompt (schematic),fonttitle=\bfseries\small,boxrule=0.4pt,arc=1pt,left=3pt,right=3pt,top=2pt,bottom=2pt]
\scriptsize
\label{fig:prompt_postcheck_box}
\textbf{Inputs.} User query; history $h_{1:k-1}$; tool schemas; prior draft $\hat{u}_k$; intervened response $\tilde{u}_k$; evaluation criteria from $\ell$.\\[2pt]
\textbf{Body.} \textit{(i)} visibility specification (only the user query, tools, and history may justify $\tilde{u}_k$); \textit{(ii)} four rejection criteria: \emph{leak} of privileged keywords, \emph{fabricated tool result}, \emph{wrong tool arguments} contradicting history or violating explicitly specified parameter values, \emph{unreasonable intervention}; \textit{(iii)} an enumerated allow-list of legitimate behaviors.\\[2pt]
\textbf{Output.} JSON $\{\texttt{ok},\,\texttt{reasons},\,\texttt{diagnosis}\}$.
\end{tcolorbox}
The verifier closes the intervention loop by ensuring that accepted trajectories are aligned with the gold path, do not leak privileged information, and stay behaviorally close to the original policy.

\subsection{Perplexity-Based Distribution Gating}
\label{sec:ppl_gating}

Although the intervention verifier removes trajectories with information leakage or implausible reasoning, the accepted trajectories may still deviate substantially from the rollout policy. When an intervened trajectory has low likelihood under the current policy, its probability ratios may fall outside the PPO clipping range, suppressing useful learning signals; meanwhile, learning from such low-probability behavior can produce poorly conditioned, high-variance gradient estimates and destabilize optimization. We therefore introduce \emph{perplexity-based distribution gating} to filter intervention trajectories with excessive off-policy deviation before policy optimization.

Specifically, after obtaining an intervened response, we remove all intervention-specific information and rescore it under the rollout policy conditioned solely on the original observable history. For $\tilde{u}_k = (\tilde{u}_k^1, \dots, \tilde{u}_k^{T_k})$ with $T_k$ tokens at step $k$, its perplexity is computed as
\begin{equation}
\label{eq:ppl_score}
\operatorname{PPL}_{\mathrm{roll}}(\tilde{u}_k)=
\exp\left(
-\frac{1}{T_k}
\sum_{t=1}^{T_k}
\log \pi_{\theta}^{\mathrm{roll}}
\left(
\tilde{u}_k^t
\mid
h_{1:k-1},\tilde{u}_k^{<t}
\right)
\right),
\end{equation}
where $T_k$ is the number of tokens in the intervened response. Higher perplexity indicates a larger deviation from the behavior supported by the policy model. 
We retain $\tilde{u}_k$ with a threshold $\delta$:
\begin{equation}
\label{eq:ppl_gate}
\operatorname{PPL}_{\mathrm{roll}}(\tilde{u}_k)
\leq \delta.
\end{equation}


\subsection{Reasoning-Shift-Aware Token Reweighting}
\label{sec:reweighting}

The intervention replaces a single failed rollout in the group with an intervened trajectory $\tilde{\tau}$, which is sampled from a behavior distribution $\pi^{\mathrm{cal}}$ different from the rollout distribution $\pi_\theta^{\mathrm{roll}}$. 
Importantly, not all corrected tokens contribute equally to learning: in many cases, a small subset of tokens is critical for changing the reasoning direction and turning a failed trajectory into a successful one. We therefore introduce a \emph{reasoning-shift-aware} token reweighting, which focuses the update on those intervention-induced tokens that carry the key reasoning correction.

Consider an intervened response segment $\tilde{u}_k = (\tilde{u}_k^1, \dots, \tilde{u}_k^{T_k})$ with $T_k$ tokens at step $k$. 
We define a token-level reweighting coefficient based on the likelihood ratio between the two conditions:
\begin{equation}
\label{eq:rs_ratio}
    \rho^t = \exp\!\left(\log \pi^{\mathrm{cal}}(\tilde{u}_k^t \mid h_{1:k-1}, \tilde{u}_k^{<t}) - \log \pi_\theta^{\mathrm{roll}}(\tilde{u}_k^t \mid h_{1:k-1}, \tilde{u}_k^{<t})\right).
\end{equation}
where $\pi^{\mathrm{cal}}$ and $\pi_\theta^{\mathrm{roll}}$ denote the token likelihoods under the correction prompt $p_{\mathrm{cal}}$ (which additionally includes the validator feedback) and the rollout prompt $p_{\mathrm{roll}}$, respectively.

A larger ratio means that the token is much more characteristic of the corrected reasoning path than of the original rollout path and should therefore receive greater emphasis during learning. To avoid unstable updates caused by extreme ratios, we apply a two-sided clip around $1$ so that the ratio acts as a bounded per-token reweighting coefficient:
\begin{equation}
\label{eq:rs_clip}
    \tilde{\rho}_{\mathrm{cal}}^t = \mathrm{clip}\!\left(\rho^t,\; 1 - \epsilon_{\mathrm{cal}},\; 1 + \epsilon_{\mathrm{cal}}\right),
\end{equation}
where $\epsilon_{\mathrm{cal}}\in(0,1)$, treated as a stop-gradient constant that only reweights the advantage of the tokens it emphasizes. 
Note that the reweighting coefficient $\tilde{\rho}_{\mathrm{cal}}^t$ is only computed on tokens from intervened trajectories; for purely on-policy trajectories (i.e., when at least one rollout in the group already succeeds and no intervention is invoked), we set $\tilde{\rho}_{\mathrm{cal}}^t \equiv 1$, so that RTI reduces exactly to GRPO on those groups.
The final policy-gradient loss applies the reasoning-shift reweighting to the token-level GRPO clipped surrogate. Concretely, expanding $\mathcal{J}_{\mathrm{GRPO}}$ from Eq.~(\ref{eq:grpo}) yields:
\begin{equation}
\label{eq:rti_loss}
\begin{aligned}
\mathcal{J}_{\mathrm{RTI}}(\theta) = \mathbb{E}_{x,\{\tilde\tau_i\}}\!\Bigg[\frac{1}{G}\sum_{i=1}^{G}\frac{1}{|\tilde\tau_i|}\sum_{t=1}^{|\tilde\tau_i|}
\tilde{\rho}_{\mathrm{cal}}^t
\min\!\Big(\rho_{\mathrm{grpo}, i}^t \,A_i,\;\mathrm{clip}(\rho_{\mathrm{grpo}, i}^t,1\!-\!\epsilon_{\mathrm{clip}},1\!+\!\epsilon_{\mathrm{clip}})\,A_i\Big)\Bigg].
\end{aligned}
\end{equation}



\section{Experiments}
\label{sec:experiments}

\subsection{Experimental Setup}
\label{sec:exp_setup}


\noindent \textbf{Benchmarks.}
We evaluate RTI across five benchmarks. First, we construct \textbf{Navigation}, a multi-waypoint navigation benchmark derived from real-world map scenarios, where agents must plan routes and complete tasks by interacting with real APIs from the Amap Open Platform. Second, we adopt the publicly available multi-turn dialogue benchmark $\tau^2$-Bench~\citep{tau2bench} and evaluate on its \textbf{Telecom}, \textbf{Retail}, and \textbf{Airline} domains. The corresponding RL training instances are drawn from the AReaL dataset~\citep{areal}. Finally, we consider Claw-Eval~\citep{ye2026claweval}, a human-verified benchmark comprising highly diverse tasks grounded in real-world environments. To evaluate whether RTI can improve performance beyond the policy's native capabilities, we select 20 challenging T-series tasks on which the base policy rarely succeeds, forming \textbf{Claw-Eval-Hard20}, and synthesize task-specific training data for these tasks. 

\noindent \textbf{Evaluation Metrics.}
We report Pass@$k$ and Pass$^k$ for $k=1,\ldots,4$. Pass@$k$ measures the proportion of tasks for which at least one of the $k$ independent rollouts succeeds, while Pass$^k$ measures the proportion of tasks for which all $k$ rollouts succeed. The p@1 is the mean Pass@$1$ over the four evaluations, while avg p@$k$ and avg p$^k$ additionally average the corresponding metric over $k=1,\ldots,4$. 
For the publicly available benchmarks, we follow their original evaluation protocols to determine whether a trajectory is successful. For Navigation, a trajectory is considered successful if the agent invokes the \texttt{direction} tool with the correct parameters.

\noindent \textbf{Baselines.}
We compare RTI against: 
\textbf{Reference models} include the backbones (Qwen3-8B / Qwen3.5-9B), larger same-family variants (Qwen3-235B and Qwen3.5-397B-A17B), and contemporaneous proprietary models, paired by generation: GPT-4.1 and Claude-4-Sonnet for Qwen3-8B, and GPT-5.1 and Claude-4.6-Sonnet for Qwen3.5-9B. 
\textbf{On-policy RL baselines} include GRPO~\citep{shao2024deepseekmath}, GSPO~\citep{gspo}, and AEPO~\citep{aepo}. 
\textbf{Teacher-guided methods} include OPD~\citep{lu2025onpolicydistillation} and LUFFY~\citep{luffy}. 

Further details on the benchmarks and baselines are provided in Appendices~\ref{app:benchmark_details} and \ref{app:baseline}.

\subsection{Main Results}
\label{sec:main_results}

Tables~\ref{tab:main_results_qwen3} and~\ref{tab:main_results_qwen35} show that RTI consistently improves agent performance, with the largest gains on tasks where the backbone and existing RL methods struggle. Three findings stand out.

\emph{First, RTI overcomes bottlenecks in domain-specific interaction.}
With Qwen3-8B, RTI improves Pass@1 over GRPO by 24.56 points on $\tau^2$-Telecom and 22.00 points on Navigation. Telecom requires the agent to guide users through actions it cannot perform itself; Navigation requires it to recognize noisy POI results and refine its search before requesting directions. Corrected rollouts expose the policy to successful strategies for both failure modes (Appendix~\ref{app:benchmark_details}).

\emph{Second, updating trajectories with a teacher does not necessarily overcome the teacher's capability limit.}
OPD and LUFFY provide additional learning signals but achieve only 26.25\% and 30.00\% Pass@1, respectively, on Claw-Eval-Hard20 with Qwen3-8B. RTI reaches 50.00\%, indicating that its gains cannot be explained solely by supplying trajectories beyond unaided on-policy exploration.

\emph{Third, RTI extends these gains to diverse hard tasks and stronger backbones.}
On Claw-Eval-Hard20, RTI exceeds GRPO by 27.50 points with Qwen3-8B. Although Qwen3.5-9B already performs strongly on several other benchmarks, its baseline Pass@1 on this subset is 26.25\%; RTI raises it to 85.00\%, 23.75 points above the strongest competing method, LUFFY. These results suggest that RTI expands the set of tasks the policy can solve across different backbone capability levels.

\begin{table}[t]
\centering
\footnotesize
\setlength{\tabcolsep}{1.5pt}
\renewcommand{\arraystretch}{1.1}
\caption{Main results on five benchmarks with the Qwen3-8B backbone.}
\label{tab:main_results_qwen3}
\resizebox{\textwidth}{!}{
\begin{tabular}{l @{\hskip 3pt} ccc @{\hskip 3pt} ccc @{\hskip 3pt} ccc @{\hskip 3pt} ccc @{\hskip 3pt} ccc}
\toprule
& \multicolumn{3}{c|}{Navigation} & \multicolumn{3}{c|}{$\tau^2$-Telecom} & \multicolumn{3}{c|}{$\tau^2$-Retail} & \multicolumn{3}{c|}{$\tau^2$-Airline} & \multicolumn{3}{c}{Claw-Eval-Hard20} \\
\cmidrule(lr){2-4} \cmidrule(lr){5-7} \cmidrule(lr){8-10} \cmidrule(lr){11-13} \cmidrule(lr){14-16}
Method & p@1 & avg p@$k$ & avg p$^k$ & p@1 & avg p@$k$ & avg p$^k$ & p@1 & avg p@$k$ & avg p$^k$ & p@1 & avg p@$k$ & avg p$^k$ & p@1 & avg p@$k$ & avg p$^k$ \\
\midrule
Qwen3-8B & 22.67 & 28.50 & 14.17 & 36.84 & 57.02 & 21.49 & 45.61 & 71.93 & 35.53 & 31.00 & 50.50 & 16.00 & 18.75 & 24.38 & 15.94 \\
Qwen3-235B-A22B & 28.00 & 35.50 & 16.67 & 44.74 & 64.04 & 28.07 & 64.04 & 79.61 & 47.15 & 50.50 & 64.50 & 32.00 & 20.00 & 20.00 & 12.50 \\
GPT-4.1 & 23.33 & 31.33 & 18.67 & 39.47 & 58.99 & 23.68 & 71.93 & 83.55 & 52.19 & 53.00 & 66.50 & \underline{34.00} & 30.00 & 30.00 & 18.75 \\
Claude-4-Sonnet & 26.00 & 32.83 & 20.67 & 49.12 & 67.76 & 32.02 & \textbf{74.56} & \textbf{85.09} & \underline{53.95} & \underline{54.00} & \underline{68.50} & \textbf{35.50} & \underline{40.00} & \underline{46.25} & \underline{30.00} \\
\midrule
GRPO & 35.33 & 40.00 & 29.00 & 36.84 & 55.48 & 21.71 & 67.54 & 80.92 & 50.22 & 46.50 & 57.50 & 26.50 & 22.50 & 24.17 & 20.83 \\
GSPO & 28.00 & 34.50 & 18.00 & 44.96 & 64.48 & 30.66 & 63.16 & 74.78 & 46.27 & 39.00 & 55.50 & 22.50 & 26.25 & 31.67 & 19.90 \\
AEPO & 34.00 & 37.67 & 26.17 & 45.68 & 62.41 & 29.23 & 63.16 & 81.14 & 49.78 & 51.00 & 65.50 & 33.50 & 20.00 & 27.50 & 16.25 \\
OPD & \underline{37.33} & 41.50 & \underline{32.00} & \underline{55.26} & 68.64 & \underline{36.62} & 68.42 & 82.24 & 46.71 & 47.50 & 56.50 & 30.50 & 26.25 & 31.46 & 21.98 \\
\midrule
LUFFY & 36.00 & \underline{51.50} & 26.67 & 49.12 & \underline{72.59} & 27.19 & 71.93 & \underline{84.87} & 49.34 & 46.00 & 59.50 & 26.50 & 30.00 & 32.81 & 27.19 \\
\midrule
\rowcolor{blue!12}
RTI & \textbf{57.33} & \textbf{62.33} & \textbf{46.67} & \textbf{61.40} & \textbf{81.14} & \textbf{44.30} & \underline{73.68} & \textbf{85.09} & \textbf{55.92} & \textbf{56.00} & \textbf{70.00} & \textbf{35.50} & \textbf{50.00} & \textbf{62.29} & \textbf{37.71} \\
\bottomrule
\end{tabular}
}
\end{table}

\begin{table}[t]
\centering
\footnotesize
\setlength{\tabcolsep}{1.5pt}
\renewcommand{\arraystretch}{1.1}
\caption{Main results on five benchmarks with the Qwen3.5-9B backbone.}
\label{tab:main_results_qwen35}
\resizebox{\textwidth}{!}{
\begin{tabular}{l @{\hskip 3pt} ccc @{\hskip 3pt} ccc @{\hskip 3pt} ccc @{\hskip 3pt} ccc @{\hskip 3pt} ccc}
\toprule
& \multicolumn{3}{c|}{Navigation} & \multicolumn{3}{c|}{$\tau^2$-Telecom} & \multicolumn{3}{c|}{$\tau^2$-Retail} & \multicolumn{3}{c|}{$\tau^2$-Airline} & \multicolumn{3}{c}{Claw-Eval-Hard20} \\
\cmidrule(lr){2-4} \cmidrule(lr){5-7} \cmidrule(lr){8-10} \cmidrule(lr){11-13} \cmidrule(lr){14-16}
Method & p@1 & avg p@$k$ & avg p$^k$ & p@1 & avg p@$k$ & avg p$^k$ & p@1 & avg p@$k$ & avg p$^k$ & p@1 & avg p@$k$ & avg p$^k$ & p@1 & avg p@$k$ & avg p$^k$ \\
\midrule
Qwen3.5-9B & 39.33 & 53.00 & 26.17 & 92.11 & 96.49 & 89.69 & 82.46 & 93.64 & 71.05 & 60.50 & 73.00 & 47.00 & 26.25 & 31.67 & 19.90 \\
Qwen3.5-397B-A17B & 44.00 & 53.67 & 38.00 & 95.61 & 98.25 & 93.64 & 84.21 & 94.74 & 73.90 & 63.00 & 77.00 & 51.00 & 30.00 & 40.00 & 26.25 \\
GPT-5.1 & 42.00 & 54.00 & 30.17 & 95.61 & 98.03 & 93.64 & 77.19 & 92.11 & 62.28 & 66.50 & 79.00 & 53.00 & 40.00 & 47.50 & 40.00 \\
Claude-4.6-Sonnet & 58.00 & 64.17 & 53.83 & \textbf{97.37} & \textbf{98.90} & \textbf{96.05} & \textbf{91.23} & \textbf{96.93} & \textbf{85.53} & \underline{80.00} & \textbf{88.50} & \textbf{71.50} & 51.25 & 59.27 & 44.17 \\
\midrule
GRPO & \underline{71.33} & 74.83 & \underline{66.33} & 95.61 & 97.59 & 93.64 & 84.21 & 94.74 & 73.90 & 75.00 & 82.50 & 63.00 & 52.50 & 57.71 & 45.42 \\
GSPO & 68.67 & 72.33 & 63.17 & 95.61 & 98.03 & 94.30 & 85.09 & 94.96 & 75.44 & 79.00 & 85.50 & 68.00 & 51.25 & 58.75 & 44.69 \\
AEPO & 70.67 & \underline{77.17} & 64.83 & \underline{96.49} & 98.47 & 94.52 & 85.96 & 95.18 & 76.75 & 78.50 & 86.50 & 70.00 & 51.25 & 56.66 & 44.90 \\
OPD & 70.00 & 75.00 & 58.67 & \underline{96.49} & \underline{98.68} & \underline{95.18} & 86.84 & 95.61 & 78.29 & 79.50 & 84.50 & 65.00 & 51.25 & 55.94 & 47.50 \\
\midrule
LUFFY & \underline{71.33} & 76.17 & 65.17 & \underline{96.49} & \underline{98.68} & \underline{95.18} & 85.96 & 95.18 & 76.75 & 78.00 & 86.50 & 67.00 & \underline{61.25} & \underline{66.67} & \underline{54.90} \\
\midrule
\rowcolor{blue!12}
RTI & \textbf{74.67} & \textbf{78.50} & \textbf{68.67} & \underline{96.49} & \underline{98.68} & \underline{95.18} & \underline{88.60} & \underline{96.05} & \underline{81.14} & \textbf{82.00} & \underline{87.50} & \underline{70.50} & \textbf{85.00} & \textbf{94.17} & \textbf{73.96} \\
\bottomrule
\end{tabular}
}
\end{table}

\subsection{Ablation Studies}
\label{sec:ablation}

\begin{table}[t]
\centering
\footnotesize
\setlength{\tabcolsep}{2pt}
\renewcommand{\arraystretch}{1.1}
\caption{Ablation study on Navigation and $\tau^2$-Telecom with the Qwen3-8B backbone.}
\label{tab:ablation}
\resizebox{\textwidth}{!}{
\begin{tabular}{l @{\hskip 3pt} cccccccc @{\hskip 3pt} cccccccc}
\toprule
& \multicolumn{8}{c|}{Navigation} & \multicolumn{8}{c}{$\tau^2$-Telecom} \\
\cmidrule(lr){2-9} \cmidrule(lr){10-17}
Method & \textsc{pass@1} & \textsc{pass@2} & \textsc{pass@3} & \textsc{pass@4} & \textsc{pass$^{1}$} & \textsc{pass$^{2}$} & \textsc{pass$^{3}$} & \textsc{pass$^{4}$} & \textsc{pass@1} & \textsc{pass@2} & \textsc{pass@3} & \textsc{pass@4} & \textsc{pass$^{1}$} & \textsc{pass$^{2}$} & \textsc{pass$^{3}$} & \textsc{pass$^{4}$} \\
\midrule
GRPO & 35.33 & 40.00 & 41.33 & 43.33 & 35.33 & 29.33 & 26.67 & 24.67 & 36.84 & 52.63 & 63.16 & 69.30 & 36.84 & 21.93 & 17.54 & 10.53 \\
\midrule
RTI w/o Post-Check & 26.00 & 33.33 & 36.00 & 38.00 & 26.00 & 14.00 & 9.33 & 7.33 & 38.60 & 53.51 & 59.65 & 67.54 & 38.60 & 24.56 & 16.67 & 14.91 \\
RTI w/o PPL Gating & 50.00 & \underline{60.00} & \underline{64.00} & \underline{65.33} & 50.00 & 37.33 & 29.33 & 24.00 & 54.39 & 68.42 & 75.44 & 79.82 & 54.39 & 27.19 & 21.05 & 18.42 \\
RTI w/o Reweight & \underline{55.33} & 58.67 & 60.67 & 62.00 & \underline{55.33} & \underline{45.33} & \underline{42.00} & \underline{33.33} & \underline{60.53} & \underline{79.82} & \underline{85.96} & \underline{88.60} & \underline{60.53} & \underline{38.60} & \underline{30.70} & \underline{20.18} \\
\midrule
\rowcolor{blue!12}
RTI & \textbf{57.33} & \textbf{61.33} & \textbf{64.67} & \textbf{66.00} & \textbf{57.33} & \textbf{48.00} & \textbf{42.00} & \textbf{39.33} & \textbf{61.40} & \textbf{83.33} & \textbf{87.72} & \textbf{92.11} & \textbf{61.40} & \textbf{46.49} & \textbf{37.72} & \textbf{31.58} \\
\bottomrule
\end{tabular}
}
\end{table}


To isolate each component's contribution, we conduct ablation studies on Navigation and $\tau^2$-Telecom with Qwen3-8B, evaluating three variants: \textit{RTI w/o Post-Check}, which removes the \emph{Integrity-Constrained Intervention Verifier}; \textit{RTI w/o PPL Gating}, which removes the \emph{PPL-Based Distribution Gating}; and \textit{RTI w/o Reweight}, which removes the \emph{Reasoning-Shift-Aware Token Reweighting}.

\noindent \textbf{Effect of Post-Check.}
Removing the Post-Check leads to a large degradation on both benchmarks. On $\tau^2$-Telecom, the variant loses almost all of RTI's gains and falls back to the GRPO level; on Navigation, it even drops below GRPO (26.00 vs.\ 35.33 in Pass@1). Without integrity verification, the admitted interventions are frequently leak-contaminated or otherwise unlearnable, and the corrupted supervision can actively harm the policy rather than merely fail to help. 
A representative failure case is given in Appendix~\ref{app:deleak_case}.

\noindent \textbf{Effect of PPL Gating.}
Removing the PPL Gate yields the second-largest degradation on $\tau^2$-Telecom: pass@1 decreases from 61.40 to 54.39, and the all-of-$k$ metrics drop even more sharply (pass$^2$: 46.49$\rightarrow$27.19, pass$^4$: 31.58$\rightarrow$18.42). This confirms that bounding the distributional shift of intervened responses is essential: interventions that deviate excessively from the rollout policy provide unstable learning signals, so filtering them out preserves consistent optimization.

\noindent \textbf{Effect of Reweight.}
Removing the token reweighting results in the mildest overall change, yet the degradation is systematic: RTI is at least as good on all sixteen metrics, with the largest gaps on the all-of-$k$ metrics. This suggests that reweighting mainly sharpens credit assignment toward the decisive intervention tokens, improving the stability with which the corrected reasoning is reproduced across repeated attempts, whereas the exploration breadth is already supplied by the intervention. 

\subsection{Hyperparameter Study}
\label{sec:param_study}

We further study the gating threshold $\delta$ in the PPL-based distribution gating (Section~\ref{sec:ppl_gating}, Eq.~(\ref{eq:ppl_gate})), which controls the maximum deviation from the rollout policy that an intervened response may exhibit before being discarded. We sweep $\delta\in\{0, 3, 5, 20, +\infty\}$ on Navigation with the Qwen3-8B backbone. At one extreme ($\delta=0$), the gate rejects all interventions, so training falls back to the no-intervention limit; at the other ($\delta=+\infty$), no filtering is applied at all.

As shown in Figure~\ref{fig:delta_curve}, the two extremes behave as expected: with $\delta=0$, training degenerates into the GRPO run and the success rate stays nearly flat, whereas $\delta=+\infty$ improves steadily but converges to a clearly lower level. Performance improves as $\delta$ grows from $3$ to $5$ and degrades at $\delta=20$: a threshold that is too small discards too many corrective trajectories to sustain late-stage gains, while a threshold that is too large admits severely shifted interventions, whose low-probability, high-variance updates destabilize optimization---the $\delta=20$ curve indeed exhibits the highest volatility among all runs. The best trade-off is achieved at $\delta=5$, which preserves informative corrections while filtering out excessively deviated ones: it tracks $\delta=3$ closely in early training, pulls ahead in later stages, and converges to the highest final success rate. We therefore adopt $\delta=5$ in all experiments.

\subsection{Further Analysis}
\label{sec:further_analysis}


\begin{figure}[t]
    \centering
    \begin{minipage}[t]{0.48\textwidth}
        \centering
        \includegraphics[width=\linewidth]{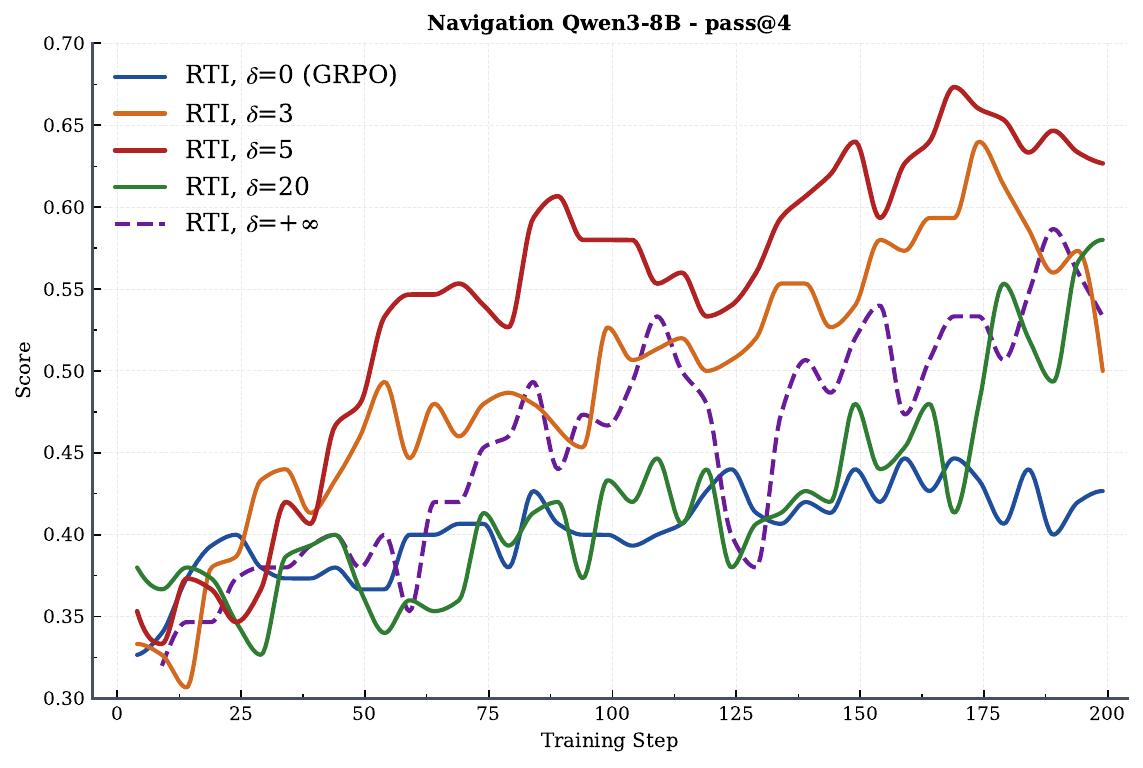}
        \vspace{-3mm}
        \captionof{figure}{Effect of the gating threshold $\delta$ on Pass@4 for Navigation with Qwen3-8B.}
        \vspace{-3mm}
        \label{fig:delta_curve}
    \end{minipage}
    \hfill
    \begin{minipage}[t]{0.48\textwidth}
        \centering
        \includegraphics[width=\linewidth]{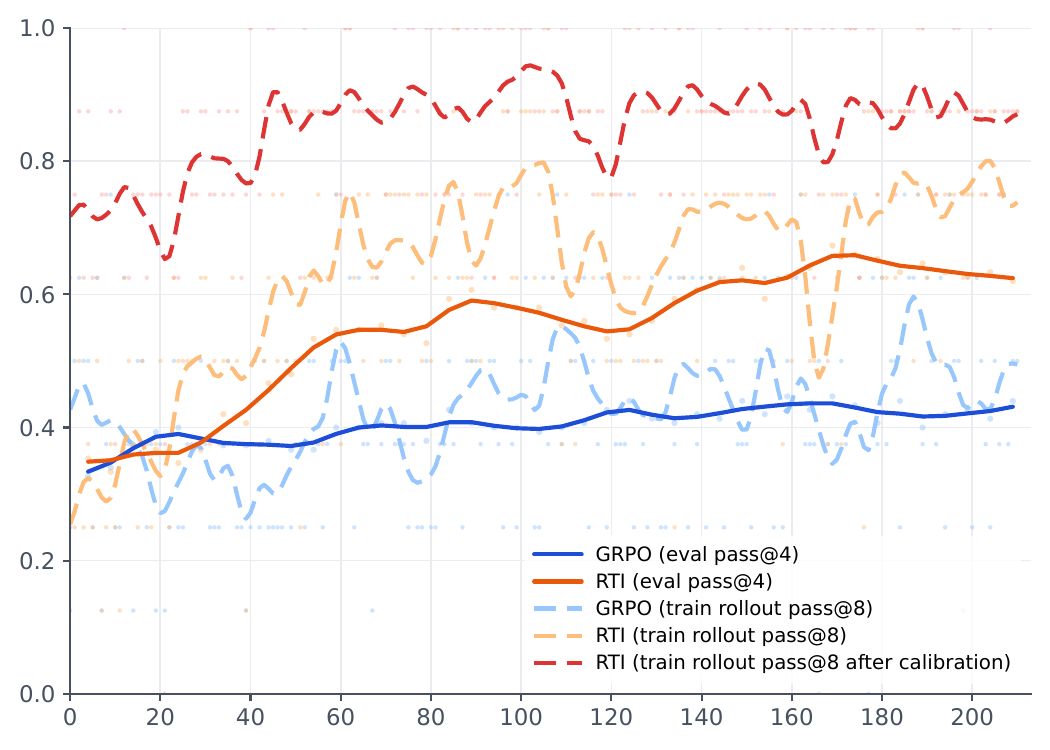}
        \vspace{-3mm}
        \captionof{figure}{Training dynamics on Navigation under GRPO vs.\ RTI with the Qwen3-8B backbone.}
        \vspace{-3mm}
        \label{fig:rl_success}
    \end{minipage}
\end{figure}

\noindent \textbf{Training Dynamics Comparison.}
Figure~\ref{fig:rl_success} compares evaluation Pass@4 (solid lines) with training rollout Pass@8 (dashed lines) on Navigation. All curves are Gaussian-smoothed for visualization. For RTI, the two dashed curves measure rollout success before and after intervention. The large gap at early stage between them shows that intervention converts many failed rollout groups into successful trajectories, providing training signals where GRPO receives few. Crucially, RTI's pre-intervention Pass@8 also rises over training and exceeds GRPO's rollout success, while its evaluation Pass@4 improves in tandem. 
Thus, the corrected trajectories support effective policy updates: the model increasingly succeeds without intervention, rather than relying on intervention only at rollout time.

\begin{figure}[t]
    \centering
    \begin{minipage}[t]{0.48\textwidth}
        \centering
        \includegraphics[width=\linewidth]{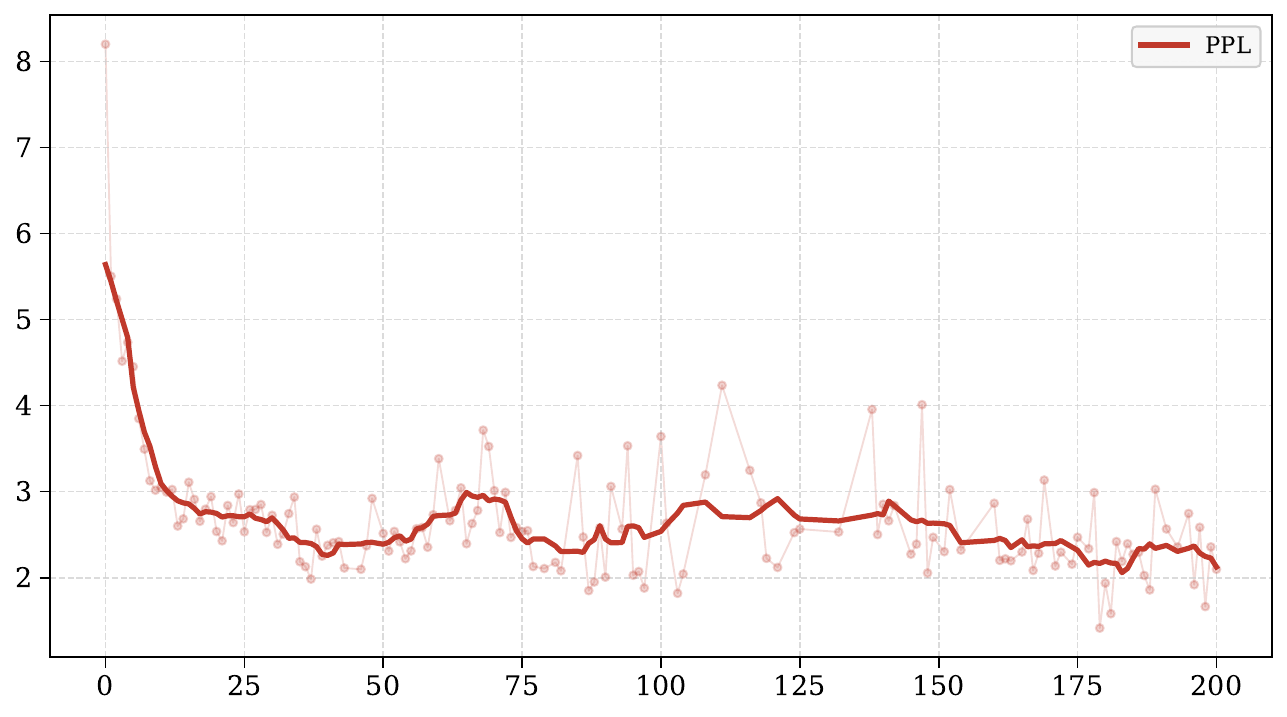}
        \subcaption{Dynamics of Intervention PPL under RTI.}
        \label{fig:inspector_ppl}
    \end{minipage}
    \hfill
    \begin{minipage}[t]{0.48\textwidth}
        \centering
        \includegraphics[width=\linewidth]{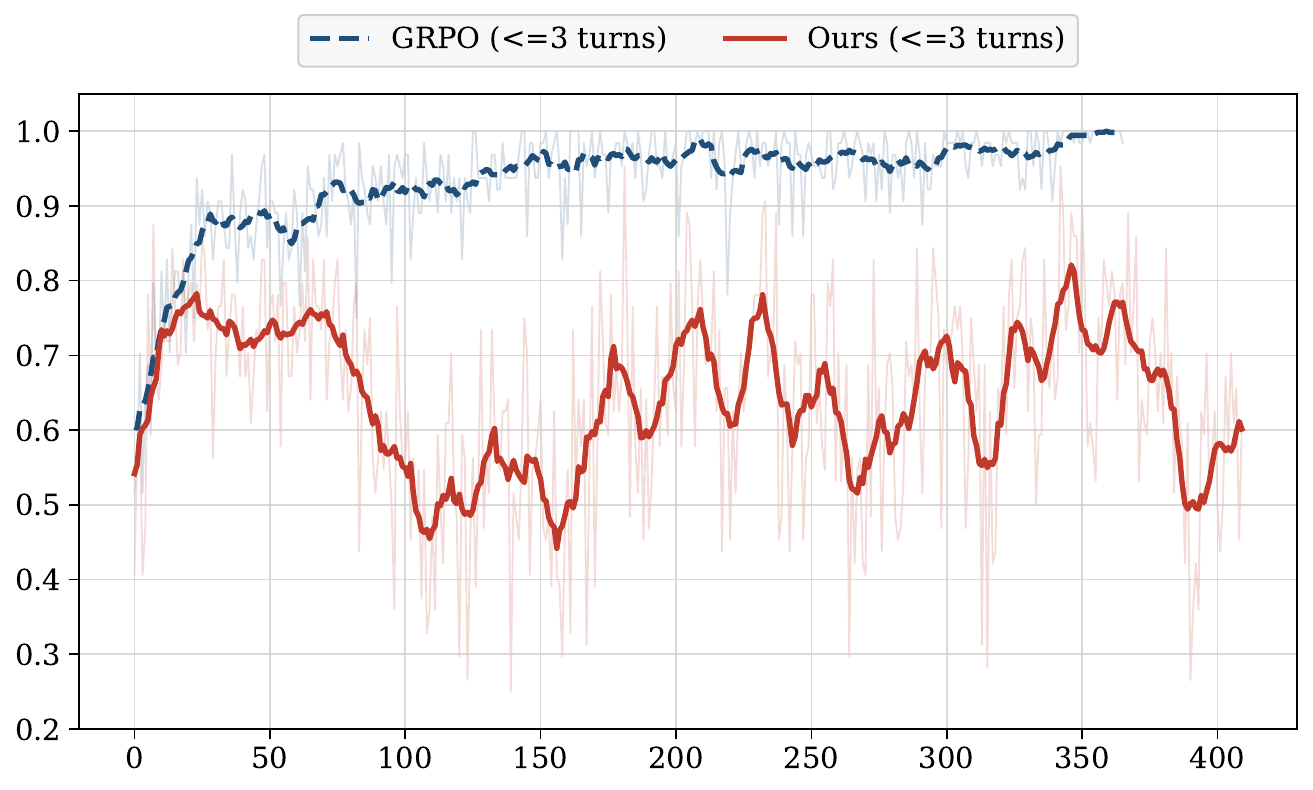}
        \subcaption{Per-iteration distribution of interaction turns.}
        \label{fig:turn_num_compare}
    \end{minipage}
    \vspace{-2mm}
    \caption{Further analysis of intervention PPL dynamics and resulting reasoning-pattern distributions.}
    \vspace{-2mm}
    \label{fig:analysis}
\end{figure}

\noindent \textbf{Dynamics of Intervened-Step Perplexity and Reasoning Patterns.} 
To further verify whether the intervened behaviors are truly internalized by the policy, we examine the average PPL under the policy model per successful intervened step, which measures the distributional gap between the intervened trajectory and the policy's native generation distribution.
As shown in Figure~\ref{fig:inspector_ppl}, the curve drops steadily during training: those that are still triggered yield intervened responses that are increasingly close to the policy's own output distribution. It provides direct evidence that the policy is not merely \emph{exploiting} the intervention at training time but is actually \emph{absorbing} the intervened reasoning pattern into its native behavior.
Additionally, we analyze Navigation to examine how training under RTI versus GRPO changes the model's \emph{resulting reasoning patterns}.
Figure~\ref{fig:turn_num_compare} contrasts the per-iteration distribution of interaction turns under GRPO and RTI on Navigation. 
Compared with GRPO, RTI maintains a broader and more multi-modal turn distribution throughout training. 
In particular, trajectories exceeding 3 turns continue to retain non-trivial mass even at later stages, indicating that the policy preserves the capacity for self-reflective re-querying rather than collapsing into a single short interaction template. A case is provided in Appendix~\ref{app:nav_case}.

\begin{figure}[t]
    \centering
    \includegraphics[width=\textwidth]{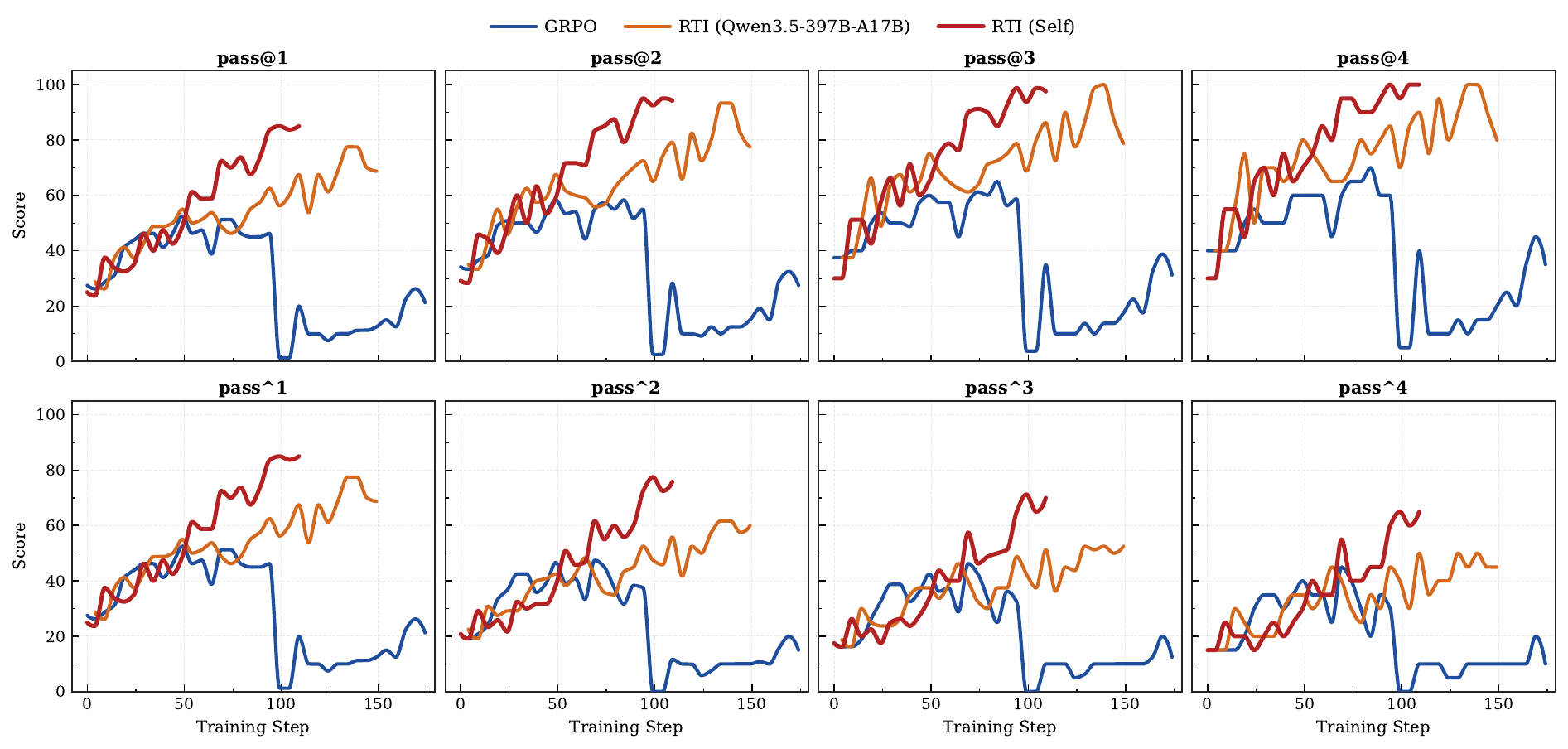}
    \caption{Training curves on Claw-Eval-Hard20 with the Qwen3.5-9B backbone.}
    \vspace{-2mm}
    \label{fig:claw-eval_curve}
    \vspace{-2mm}
\end{figure}

\noindent \textbf{Generalization Across Hard Agent Tasks.}
All preceding RTI experiments use self-intervention. To examine whether a stronger intervening model improves learning, we additionally use Qwen3.5-397B-A17B to correct rollouts from the Qwen3.5-9B policy. Figure~\ref{fig:claw-eval_curve} compares this variant with standard RTI and GRPO on Claw-Eval-Hard20. GRPO improves briefly before its Pass@$k$ and Pass$^k$ scores collapse. Both RTI variants instead sustain gains across the two metric families: higher Pass@$k$ reflects success on more tasks within $k$ attempts, while higher Pass$^k$ reflects more consistent success across attempts. Standard RTI achieves the strongest results, showing that corrections from the much larger model deviate further from the policy's distribution and are therefore harder to learn from, despite the teacher's greater capability. These findings show that self-intervention suffices to improve performance across diverse hard tasks without relying on a stronger teacher. Additionally, we supplement the experimental results on the full set of unimodal (text-only) tasks in Appendix~\ref{app:ood}.

\section{Conclusion}

We introduced RTI to enable agent policies to learn beyond their unaided exploration capacity. RTI uses reward-aware self-intervention to correct failed rollouts, a perplexity-based gate to filter excessively shifted trajectories, and reasoning-shift-aware token reweighting to focus policy updates on consequential decisions. Across five benchmarks and two backbones, RTI substantially improves task success, with particularly large gains on Navigation, $\tau^2$-Telecom, and Claw-Eval-Hard20. Training analyses show that intervention supplies successful trajectories early in training and that the policy increasingly succeeds without intervention as learning proceeds. An additional comparison using a much larger intervening model confirms that RTI does not require a stronger teacher to achieve these gains, supporting its applicability across diverse agent tasks.

\section*{Acknowledgments}

This work was supported by Alibaba Research Intern Program. We would like to thank the Qwen Team at Alibaba Token Hub (ATH), Alibaba Group, for providing the computational resources and foundation models (Qwen) used in this research.

\section*{AI use statement}
We used generative AI tools as an assistive aid in the preparation of this manuscript, specifically for proofreading, grammar checking, and language refinement to improve clarity and readability. All research ideas, experimental designs, theoretical claims, and final content were conceived, verified, and approved by the authors.

\section*{Ethics statement}

We will abide by the laws, rules, and regulations of our community, school, work, and country. We will conduct ourselves with integrity, fidelity, and honesty. We will openly take responsibility for our actions and only make agreements that we intend to keep. All data used in this study are intended for research purposes. No personally identifiable information (PII) was collected.

\section*{Reproducibility statement}

We have included comprehensive details to support reproducibility of this work. The complete RTI algorithm, implementation details, hyperparameters, benchmark descriptions, baseline configurations, and theoretical proofs are provided in the appendix. Representative prompts and case studies are also included in the appendix.


\newpage
\appendix

\section{More Details of RTI}

\subsection{RTI MAS Prompt Templates}
\label{app:prompts}

In this section, we present the full prompt templates for the three MAS agents used in RTI.

\subsubsection{$\tau^2$-Bench Prompts (Telecom / Retail / Airline)}

\paragraph{Reward-Oriented Trajectory Validator.}
The Validator assesses whether the current response is a valid next step that keeps the agent on track. It checks against a set of hard rules and marks the response invalid only when a clear violation is detected.

\begin{tcolorbox}[title={\textbf{Reward-Oriented Trajectory Validator ($\tau^2$-Bench)}},colback=gray!4,colframe=gray!50,boxrule=0.4pt,arc=1pt,left=3pt,right=3pt,top=3pt,bottom=3pt,breakable,fonttitle=\scriptsize]
\scriptsize
\texttt{You are a trajectory supervisor for a tool-calling agent task. Your job is to judge whether the CURRENT response is a valid next step that keeps the agent on track toward completing the task. Task is complete only when \{completion\_note\}.}

\texttt{[Observable History] \{history\_json\}}

\texttt{[Current Response](Formatted as json) \{response\_text\}}

\texttt{[Required actions to complete the task]}\\
\texttt{Agent-side actions (the agent must call these tools itself): \{agent\_actions\}}\\
\texttt{\{user\_actions\_block\}}~~\textit{(telecom only: lists user-side device actions the agent must guide)}

\texttt{How to tell if an action is done:}\\
\texttt{- agent-side action: a matching assistant tool call + successful result in history.}\\
\texttt{- user-side action [telecom]: a matching user tool call output in history.}

\texttt{RULES --- mark invalid if ANY rule is violated:}\\
\texttt{1) FABRICATED\_ARGUMENTS: A tool argument value is not present in history and not derivable from the task description. (Values from earlier tool outputs ARE valid.)}\\
\texttt{2) POINTLESS\_REPEAT: Calling the exact same tool with the exact same arguments that already succeeded. (Same tool, different arguments = allowed.)}\\
\texttt{3) FALSE\_CLAIM: The response claims information was retrieved without a corresponding tool call.}\\
\texttt{4) [domain\_specific]:}\\
\texttt{\quad [Telecom] USER\_SIDE\_DEVICE\_ACTION: The agent emits a tool call for a user-device action. The agent must NEVER call these --- it must instruct the user to perform them on their device.}\\
\texttt{\quad\quad USER\_SIDE\_TOOLS = [check\_status\_bar, check\_network\_status, ..., reboot\_device]}\\
\texttt{\quad\quad AGENT\_SIDE\_TOOLS = [get\_customer\_by\_phone, ..., refuel\_data]}\\
\texttt{\quad [Retail] MISSING\_CONFIRMATION: Before any tool call that updates the database (including modify\_user\_address and any cancel/modify/return/exchange action), the agent must first list the exact action details and obtain an explicit user confirmation ``yes''.}\\
\texttt{\quad [Airline] MISSING\_CONFIRMATION: Before taking any actions that update the booking database (booking, modifying flights, editing baggage, changing cabin class, or updating passenger information), the agent must first list the action details and obtain an explicit user confirmation ``yes''.}\\
\texttt{5) PREMATURE\_TERMINATION: The response ends the conversation (transfer to human, closing message) while one or more required actions are not yet done.}\\
\texttt{6) CONTRADICTORY\_ACTION: The response calls a tool or instructs the user in a way that directly contradicts a required action.}\\
\texttt{7) GOAL\_DRIFT: The response ignores an obviously pending required action and pursues something unrelated.}\\
\texttt{8) ARGUMENT\_MISMATCH: For actions that update the database, if [Required actions] specifies exact argument values, the agent's tool call must match these values exactly.}

\texttt{DO NOT mark invalid for:}\\
\texttt{- Suboptimal but recoverable action order}\\
\texttt{- Diagnostic / information-lookup steps before acting}\\
\texttt{- Re-attempting a failed action with corrected arguments}\\
\texttt{- A valid path that differs from the reference}

\texttt{[Evaluation\_Criteria --- for context only] \{evaluation\_criteria\}}

\texttt{Output JSON only:}\\
\texttt{\{"diagnosis": "...", "is\_valid": true/false,}\\
\texttt{~"error\_types": [...], "suggest\_next": "..."\}}
\end{tcolorbox}

\paragraph{Policy-Aligned Trajectory Corrector.}
The Corrector generates an optimized next action that corrects errors while concealing privileged information. It receives the Validator's diagnosis together with the visible-only conversation history and produces an intervened response.

\begin{tcolorbox}[title={\textbf{Policy-Aligned Trajectory Corrector ($\tau^2$-Bench)}},colback=gray!4,colframe=gray!50,boxrule=0.4pt,arc=1pt,left=3pt,right=3pt,top=3pt,bottom=3pt,breakable,fonttitle=\scriptsize]
\scriptsize
\texttt{You are a silent calibrator for a tool-calling agent task. Your role is to minimally correct the agent's next action to bring it closer to the successful path, WITHOUT exposing any privileged knowledge.}

\texttt{[Conversation History] \{history\}}

\texttt{[Available Tools] \{tools\}}

\texttt{[Reference / evaluation criteria (for calibration only; MUST NOT appear in output)] \{evaluation\_criteria\}}

\texttt{CORE INSTRUCTIONS:}\\
\texttt{Generate the optimized next action based on [Conversation History], NOT by continuing from [Prior Draft]. The Prior Draft is provided only as a style reference.}

\texttt{GENERATION STRATEGY:}\\
\texttt{1. Style alignment: match the reasoning style, verbosity, and expression habits of the Prior Draft.}\\
\texttt{2. Progressive correction: guide toward a more reasonable path; corrections should feel like natural evolution of thought.}\\
\texttt{3. Concealment: NEVER mention ``reference'', ``groundtruth'', ``required actions'', ``evaluation criteria'', ``prior draft'', ``calibrated response'', ``previous version'', or ``the correct answer''. All reasoning must appear naturally derived from conversation history.}\\
\texttt{4. Minimal change: only correct clearly wrong steps; preserve correct parts of the prior draft.}

\texttt{CORRECTION GUIDELINES:}\\
\texttt{- If a tool argument is wrong, use the correct value from history (e.g., correct payment\_id from user profile)}\\
\texttt{- If the agent skipped a necessary step, add a message to gather required info or confirm with user}\\
\texttt{- If the agent made an unfounded claim, revise to acknowledge uncertainty or check first}\\
\texttt{- If multiple valid approaches exist, prefer the one closer to the original intent}\\
\texttt{- DO NOT introduce information that cannot be derived from visible history}\\
\texttt{- If no valid agent-side tool call is appropriate, set ``tool\_calls'' to an empty list.}
\texttt{- Before any tool call that updates the database (including modify\_user\_address and any cancel/modify/return/exchange action), the agent must first list the exact action details and obtain an explicit user confirmation ``yes''.}

\texttt{[domain\_specific]}\\
\texttt{[Telecom] If an action must be done by the user on their device, do not emit a tool call for it; guide the user to perform the action on their device instead.}\\
\texttt{\quad USER\_SIDE\_DEVICE\_ACTIONS = [check\_status\_bar, ..., reboot\_device]}\\
\texttt{\quad AGENT\_SIDE\_TOOLS = [get\_customer\_by\_phone, ..., refuel\_data]}\\
\texttt{\quad Note: ``tool\_calls'' must be an empty list if no agent-side tool is to be called.}\\

\texttt{[Prior Draft (style reference only)] \{original\_response\}}

\texttt{[Supervisor Feedback (problem indication)] \{feedback\_json\}}

\texttt{Output JSON: \{"reasoning": "...", "tool\_calls": [...], "content": "..."\}}\\
\texttt{Field rules: ``reasoning'' is always required. ``tool\_calls'' and ``content'' are mutually exclusive: use tool\_calls for agent-side actions, content for direct user replies.}
\end{tcolorbox}

\paragraph{Integrity-Constrained Intervention Verifier.}
The Verifier validates that the corrected response does not leak privileged information or introduce hallucinated content. It serves as the final quality gate before an intervened response enters the training batch.

\begin{tcolorbox}[title={\textbf{Integrity-Constrained Intervention Verifier ($\tau^2$-Bench)}},colback=gray!4,colframe=gray!50,boxrule=0.4pt,arc=1pt,left=3pt,right=3pt,top=3pt,bottom=3pt,breakable,fonttitle=\scriptsize]
\scriptsize
\texttt{You are a post-calibration safety \& faithfulness inspector. Decide whether the CALIBRATED reply should be accepted.}

\texttt{VISIBLE INFO = user question + tool list + conversation history (including tool outputs). Nothing else is allowed.}

\texttt{[User Question] \{user\_prompt\}}\\
\texttt{[Evaluation Criteria] \{evaluation\_criteria\}}\\
\texttt{[Available Tools] \{tools\}}\\
\texttt{[Agent History (visible)] \{history\}}\\
\texttt{[Prior Draft] \{original\}}\\
\texttt{[Calibrated Reply] \{corrected\}}

\texttt{Reject (ok=false) ONLY if ANY holds:}\\
\texttt{1) leak: explicitly mentions ``reference'', ``groundtruth'', ``evaluation criteria'', ``required actions'', ``answer key'', ``labels'', ``seeing the solution'', ``the prior draft'', ``previous version'', ``calibrated version'', or any phrasing that implies the model is comparing its output to a hidden draft or ground truth.}\\
\texttt{2) fabricated\_tool\_result: claims a specific tool result/value that does NOT appear in history. Example: claiming ``flight price is \$500'' when no tool returned that exact value.}\\
\texttt{3) wrong\_tool\_arguments: uses tool arguments that contradict visible history or deviate from explicit constraints defined in [Evaluation Criteria]. Mandatory Alignment: if the criteria explicitly specifies a parameter value, the agent's argument must match exactly. Contextual Discretion: if not specified, the agent may derive values logically.}\\
\texttt{4) unreasonable\_calibration: the calibrated reply is not a reasonable next step given the visible history, fails to improve the prior draft, or introduces unnecessary or less appropriate actions.}

\texttt{ALLOWED (not an issue) --- DO NOT REJECT for:}\\
\texttt{- Adding reasonable next steps implied by task context (e.g., asking for user confirmation before database updates)}\\
\texttt{- Adding policy-compliant behaviors (e.g., ``I need to verify with the user first'')}\\
\texttt{- Changing the action sequence to be more cautious or thorough}\\
\texttt{- Adding intermediate steps that logically follow from the conversation}

\texttt{The calibration goal is to IMPROVE the agent's behavior, not just paraphrase it. Adding missing confirmation steps or policy-compliant actions is the INTENDED behavior of calibration.}

\texttt{Output JSON: \{"ok": true/false, "reasons": [...], "diagnosis": "..."\}}
\end{tcolorbox}

\subsubsection{Navigation Domain Prompts}

The Navigation domain follows the same three-agent Validate--Correct--Verify pipeline as $\tau^2$-Bench, sharing the identical architectural logic: the Validator detects deviations from a valid path using the reward-side evidence as a silent reference; the Corrector applies the same four generation principles (style alignment, progressive correction, leakage suppression, minimal intervention); and the Verifier enforces the same integrity constraints (no leakage, no hallucination).

\paragraph{Reward-Oriented Trajectory Validator (Navigation).}
The Navigation Validator applies the same goal-oriented, process-agnostic philosophy as its $\tau^2$-Bench counterpart: it does not penalize suboptimal-but-recoverable action orders or stylistic differences, and only flags violations of the hard rules already stated in the agent's system prompt.

\begin{tcolorbox}[title={\textbf{Reward-Oriented Trajectory Validator (Navigation)}},colback=gray!4,colframe=gray!50,boxrule=0.4pt,arc=1pt,left=3pt,right=3pt,top=3pt,bottom=3pt,breakable,fonttitle=\scriptsize]
\scriptsize
\texttt{You are a trajectory supervisor for an ``Agent tool-calling task''. Judge whether the current response constitutes a reasonable next step.}

\texttt{[Conversation History (observable evidence)] \{history\_json\}}

\texttt{[Current Response] \{response\_text\}}

\texttt{[Hard Rules]:}\\
\texttt{- No fabricating coordinates/key parameters (must come from historical tool returns)}\\
\texttt{- Do NOT call direction before all POI coordinates are collected}\\
\texttt{- No redundant poi\_search with identical parameters in the same turn}\\
\texttt{- Parallel tool calls are allowed}

\texttt{[Reference (for internal comparison only; do NOT judge invalid based solely on reference)]}\\
\texttt{order: \{order\}~~~ref: \{ref\}~~~groundtruth: \{gt\}}

\texttt{Mark invalid ONLY if:}\\
\texttt{A) Hard rule violation (fabrication / premature direction / redundant search)}\\
\texttt{B) Stubbornly repeating on clearly failed tool returns}\\
\texttt{C) Irreversible wrong critical call with clearly untenable arguments}

\texttt{Output JSON only:}\\
\texttt{\{"is\_valid": true/false, "error\_types": [],}\\
\texttt{~"diagnosis": "...(cite observable evidence/hard rules)...",}\\
\texttt{~"suggest\_next": "..."\}}
\end{tcolorbox}

\paragraph{Policy-Aligned Trajectory Corrector (Navigation).}
The Navigation Corrector shares the same four generation principles as $\tau^2$-Bench (style alignment, progressive correction, leakage suppression, minimal intervention).

\begin{tcolorbox}[title={\textbf{Policy-Aligned Trajectory Corrector (Navigation)}},colback=gray!4,colframe=gray!50,boxrule=0.4pt,arc=1pt,left=3pt,right=3pt,top=3pt,bottom=3pt,breakable,fonttitle=\scriptsize]
\scriptsize
\texttt{You are the ``silent calibrator'' in a navigation task --- your role is to minimally optimize an agent's action trajectory toward the successful path, WITHOUT exposing any additional knowledge. You have a global view (knowing how the task should ultimately be completed), but must translate this insight into seemingly natural, progressive corrections.}

\texttt{[Conversation History] \{history\}}

\texttt{[Available Tools] \{tools\}}

\texttt{[Reference/groundtruth (for calibration only; MUST NOT appear in output)]}\\
\texttt{ref: \{ref\}~~~order: \{order\}~~~groundtruth: \{gt\}}

\texttt{CORE INSTRUCTIONS: Generate the optimized next action based on [Conversation History], NOT based on [Prior Draft]. The prior draft is provided only as a style reference.}

\texttt{GENERATION STRATEGY:}\\
\texttt{1. Style alignment: analyze the reasoning style, detail level, and expression habits of the Prior Draft; new reasoning should match in style.}\\
\texttt{2. Progressive correction: leverage your global view to embed a more reasonable reasoning path; corrections should feel like natural evolution, not a 180-degree turn.}\\
\texttt{3. Concealment: NEVER mention or imply you know the correct answer, final goal, reference behavior, or full process. All reasoning must appear naturally derived from history.}\\
\texttt{4. Minimal change: only correct clearly wrong key points; preserve correct parts of the prior draft.}

\texttt{[Prior Draft (style reference)] \{original\_response\}}

\texttt{[Supervisor Feedback (problem indication)] \{feedback\_json\}}

\texttt{KEY REMINDERS:}\\
\texttt{- You are generating the next action based on [Conversation History], not continuing from [Prior Draft]}\\
\texttt{- Your output should look like the agent's own naturally improved reasoning}
\end{tcolorbox}

\paragraph{Integrity-Constrained Intervention Verifier (Navigation).}
The Navigation Verifier enforces the same two core rejection criteria as the $\tau^2$-Bench Verifier---leakage of privileged information and hallucination of underivable content.

\begin{tcolorbox}[title={\textbf{Integrity-Constrained Intervention Verifier (Navigation)}},colback=gray!4,colframe=gray!50,boxrule=0.4pt,arc=1pt,left=3pt,right=3pt,top=3pt,bottom=3pt,breakable,fonttitle=\scriptsize]
\scriptsize
\texttt{You are a ``safety \& explainability post-checker''. A calibrated reply is given; decide whether it should be accepted.}

\texttt{[User Question] \{user\_prompt\}}\\
\texttt{[Available Tools] \{tools\}}\\
\texttt{[Agent Conversation History] \{history\}}\\
\texttt{[Prior Draft] \{original\}}\\
\texttt{[Calibrated Reply] \{corrected\}}

\texttt{Rejection criteria (any one triggers ok=false):}\\
\texttt{1) leak: mentions reference/groundtruth/standard answer/labels, or clearly uses invisible information.}\\
\texttt{2) not\_derivable (hallucination): the calibrated reply adds/modifies key facts or tool call parameters that come neither from ``user question + visible history'' nor qualify as ``allowed rewrites'' below.}

\texttt{Allowed rewrites (NOT hallucination):}\\
\texttt{- Equivalent rewriting of already-appeared information (synonyms, format normalization, unit conversion)}\\
\texttt{- Incorporating already-specified context constraints into parameters (region, time window, mode)}\\
\texttt{- Conservative narrowing for disambiguation (combining category words with keywords, removing uncertain components)}\\
\texttt{- Decomposing complex requests into verifiable sub-steps (but must not claim tool results that don't exist in history)}

\texttt{Output JSON only:}\\
\texttt{\{"ok": true/false, "reasons": ["leak", "not\_derivable"],}\\
\texttt{~"diagnosis": "must cite observable evidence (specific fragments from history / missing points)"\}}
\end{tcolorbox}

\subsection{The Full RTI Algorithm}
\label{sec:rti_algorithm}

Algorithm~\ref{alg:rti} integrates the intervention pipeline and the on-policy GRPO loop into a single training procedure. At every iteration we sample a batch of queries from $\mathcal{D}$ and, for each query, first run a fully on-policy GRPO group; if any rollout in this group succeeds, the group is used as-is and only the on-policy portion of $\mathcal{J}_{\mathrm{RTI}}$ is exercised (with $\tilde\rho_{\mathrm{cal}}^t \equiv 1$). Only when the entire group fails do we re-run an intervened rollout with the MAS intervention pipeline enabled; every accepted correction must additionally pass the perplexity gate of Eq.~(\ref{eq:ppl_gate}), and only if the intervened rollout succeeds is a single failed member of the group replaced by the intervened trajectory, for which we additionally cache the rollout-prompt and correction-prompt logprobs needed by Eq.~(\ref{eq:rti_loss}). A single backward pass over $\mathcal{J}_{\mathrm{RTI}}$ then updates $\theta$ with a gradient that is on-policy on easy instances, reweighted on intervened tokens of hard instances, and unbiased on every other token. This design preserves GRPO's behavior whenever the policy can solve a task on its own and adds capability-expanding signals exactly where vanilla on-policy RL stalls.

\begin{algorithm}[t]
\small
\caption{Rollout-Time Intervention (RTI)}
\label{alg:rti}
\SetAlgoLined
\DontPrintSemicolon
\KwIn{Policy $\pi_\theta$ (which also instantiates all MAS roles), reference $\pi_{\mathrm{ref}}$, task distribution $\mathcal{D}$, group size $G$, reweighting clip $\epsilon_{\mathrm{cal}}$.}
\For{each training iteration}{
    Sample batch $\{x_n\}\sim\mathcal{D}$\;
    \For{each query $x_n$}{
        \tcp{Stage~1: on-policy rollouts}
        Sample $\{\tau_i\}_{i=1}^{G}\sim\pi_\theta(\cdot\mid x_n)$ with intervention disabled\;
        \If{$\max_i r(x_n,\tau_i)>0$}{
            $\{\tilde\tau_i\}\leftarrow\{\tau_i\}$, set $\tilde\rho_{\mathrm{cal}}^t\equiv 1$ on all tokens\tcp*{purely on-policy update}
        }
        \Else{
            \tcp{Stage~2: reward-aware MAS intervention}
            $\{\tilde\tau_i\}\leftarrow\{\tau_i\}$\;
            Run an intervened rollout: at each step $k$ generate $\hat u_k\sim\pi_\theta$;\;
            pass $(h_{1:k-1},\hat u_k,\ell)$ through Validator $\rightarrow$ Corrector $\rightarrow$ Verifier\;
            \If{the correction is accepted by the Verifier}{
                Rescore $\tilde u_k$ under the rollout prompt and compute $\operatorname{PPL}_{\mathrm{roll}}(\tilde u_k)$ by Eq.~(\ref{eq:ppl_score})\;
                \If{$\operatorname{PPL}_{\mathrm{roll}}(\tilde u_k)\leq\delta$}{
                    Advance with $\tilde u_k$, cache $\log\pi_\theta^{\mathrm{roll}}$ and $\log\pi^{\mathrm{cal}}$\;
                }
                \Else{
                    Discard $\tilde u_k$, advance with $\hat u_k$\;
                }
            }
            \Else{
                Advance with $\hat u_k$\;
            }
            \If{the intervened rollout achieves reward $1$}{
                Replace a single failed member of $\{\tilde\tau_i\}$ with the intervened trajectory\;
                Compute clipped weights $\tilde{\rho}_{\mathrm{cal}}^t$ via Eq.~(\ref{eq:rs_clip}) on intervened tokens; set $\tilde{\rho}_{\mathrm{cal}}^t\equiv 1$ on the remaining on-policy tokens\;
            }
            \Else{
                Keep $\{\tilde\tau_i\}$ unchanged, so that no correction enters the update\;
            }
        }
        Compute group advantages $\{A_i\}$ from rewards $\{r(x_n,\tilde\tau_i)\}$ as in GRPO\;
    }
    Update $\theta$ by ascending the RTI objective (Eq.~\ref{eq:rti_loss})\;
}
\end{algorithm}

\section{Theoretical Analysis of RTI}

\label{app:convergence}

\subsection{Convergence Analysis of RTI}
RTI differs from standard on-policy GRPO in two aspects: it incorporates MAS-intervened trajectories, which are off-policy relative to the current policy, and it further applies the reasoning-shift-aware token reweighting in Eq.~\eqref{eq:rti_loss}. These two components raise a natural concern: although they may improve exploration and credit assignment, they also introduce additional distribution shift into the optimization process.

Based on the theoretical analysis of stochastic gradient descent in nonconvex optimization~\citep{nonconvex_optimization,luffy}, we give a convergence analysis in Theorem~\ref{thm:convergence} to show that, under the alignment and decoupled boundedness conditions stated below, RTI admits an $O(1/\sqrt{K})$ stationarity guarantee, where $K$ is the total number of iterations. For notational simplicity, we write $\mathcal{J}(\theta) := \mathcal{J}_{\mathrm{RTI}}(\theta)$ throughout this section.

\begin{theorem}[Conditional Convergence under Alignment and Decoupled Boundedness]
\label{thm:convergence}
RTI constructs the policy gradient using a ratio mixture $\lambda \in (0,1)$ of off-policy intervened rollouts and on-policy rollouts, where $\lambda$ denotes the expected mixing ratio over training (since intervention is triggered only on fully failed groups, $\lambda$ is a data-dependent random quantity and we take its expectation as a constant for the analysis). Accordingly, we decompose the per-step update as $\Delta\theta^k = \lambda\,\Delta\theta_{\mathrm{cal}}^k + (1-\lambda)\,\Delta\theta_{\mathrm{grpo}}^k$. Suppose that the objective $\mathcal{J}(\theta)$ is $L$-smooth and that $\mathcal{J}^\star := \sup_\theta \mathcal{J}(\theta) < \infty$. Assume further that the expected update direction is aligned with the true gradient with a lower bound governed by the on-policy clip margin $1-\epsilon_{\mathrm{clip}}$ (this alignment lower bound arises from the on-policy term $\Delta\theta_{\mathrm{grpo}}^k$ under the classic clipped surrogate; the intervened term $\Delta\theta_{\mathrm{cal}}^k$ is not required to contribute alignment and is instead absorbed into the second-moment bound $D_{\mathrm{cal}}^2$ in Definition~\ref{definition_5}), i.e.,
\begin{equation}
\mathbb{E}\!\left[\left\langle \nabla \mathcal{J}(\theta^k), \Delta \theta^k \right\rangle\right]
\ge
\alpha_k (1-\epsilon_{\mathrm{clip}})\|\nabla \mathcal{J}(\theta^k)\|^2.
\end{equation}
For $K$ iterations, let the learning rate be set to $\alpha_k = c/\sqrt{K}$, where
\begin{equation}
c = \sqrt{\frac{2 \left[ \mathcal{J}^\star - \mathcal{J}(\theta^0) + \frac{L}{2} \lambda D_{\mathrm{cal}}^2 \right]}{L  (1 + \epsilon_{\mathrm{clip}})^2 \sigma^2}}    
\end{equation}
Then the iterates satisfy the following $\mathcal{O}(1/\sqrt{K})$ convergence bound:
\begin{equation}
\small
\begin{aligned}
\min_{0 \le k \le K-1} \mathbb{E}\left[\|\nabla \mathcal{J}(\theta^k)\|^2\right] \le \frac{1}{(1 - \epsilon_{\mathrm{clip}})} \sqrt{\frac{2 \left[ \mathcal{J}^\star - \mathcal{J}(\theta^0) + \frac{L}{2} \lambda D_{\mathrm{cal}}^2 \right] L  (1 + \epsilon_{\mathrm{clip}})^2 \sigma^2}{K}},
\end{aligned}
\end{equation}
where $\theta^0$ denotes the initial policy parameter, $\sigma^2$ controls the base gradient scale and $D_{\mathrm{cal}}^2 = 4\,\epsilon_{\mathrm{cal}}^2\,\eta^2\,\sigma^2$ with $\eta > 0$ as a parameter-space perturbation constant abstracting the bounded contribution of the intervened rollouts (formalized in Definition~\ref{definition_5}).
\end{theorem} 

\subsection{Proof of Theorem~\ref{thm:convergence}}

We first state the standard definitions and assumptions for non-convex stochastic optimization.

\begin{definition}[Stochastic Oracle]
\label{def:stochastic_oracle}
For a smooth functional $\mathcal{J}$, we assume the algorithm has access to a stochastic oracle that, given an index $i \in [n]$ and a parameter $\theta \in \mathbb{R}^d$, returns the pair $(\mathcal{J}_i(\theta), \nabla \mathcal{J}_i(\theta))$.
\end{definition}

\begin{definition}[Lipschitz Smoothness]
\label{def:lipschitz}
We say $\mathcal{J} : \mathbb{R}^d \to \mathbb{R}$ is $L$-smooth if there exists a constant $L > 0$ such that $\|\nabla \mathcal{J}(\vartheta) - \nabla \mathcal{J}(\theta)\| \le L\|\vartheta - \theta\|, \forall \vartheta, \theta \in \mathbb{R}^d$. This ensures the objective's curvature is bounded.
\end{definition}

\begin{definition}[$\sigma$-bounded base gradients]
\label{def:sigma_bounded}
We say $\mathcal{J}$ has $\sigma$-bounded base gradients if the unscaled stochastic gradient satisfies $\|\nabla \mathcal{J}_i(\theta)\| \leq \sigma$ for all $i \in [n]$ and $\theta \in \mathbb{R}^d$.
\end{definition}

\begin{definition}[$\epsilon$-accuracy]
\label{def:epsilon_accuracy}
A stochastic iterative algorithm is said to achieve $\epsilon_{\mathrm{acc}}$-accuracy in $K$ iterations if the expected gradient norm satisfies $\min_{0 \le k \le K-1} \mathbb{E}[\|\nabla \mathcal{J}(\theta^k)\|^2] \le \epsilon_{\mathrm{acc}}$.
\end{definition}

To ensure stable optimization, we construct the learning objective as a deterministic weighted combination of two distinct data sources using a proportion scalar $\lambda \in (0, 1)$. At any training step $k$, exactly a fraction $\lambda$ of the learning signal originates from the off-policy intervened trajectories $\mathcal{D}_{\mathrm{cal}}$, and the remaining $1-\lambda$ originates from the on-policy rollout policy $\mathcal{D}_{\mathrm{grpo}}$. 
For a given trajectory, the stochastic gradient is scaled by a token-level reweighting coefficient. We define the objective surrogates asymmetrically:

\textbf{Off-policy Intervened Rollouts:} To push the current policy beyond its existing capability boundary, we employ the reweighting $\tilde{\rho}_{\mathrm{cal}}^t$ defined in Eq.~\eqref{eq:rs_clip}, which is bounded within $[1-\epsilon_{\mathrm{cal}},\,1+\epsilon_{\mathrm{cal}}]$. 
In practice, this coefficient is computed separately from the policy-gradient term and is detached from the computation graph; it therefore acts purely as a bounded per-token weight on the advantage:
\begin{equation}
\label{eq_pos}
\tilde{r}^t_{\mathrm{cal}} = \text{sg}\left(\tilde{\rho}_{\mathrm{cal}}^t\right),
\end{equation}
where $\text{sg}(\cdot)$ denotes the stop-gradient operator. Since the coefficient is not differentiated, it introduces no additional gradient path through $\pi^{\mathrm{cal}}$ or $\pi_\theta^{\mathrm{roll}}$: it preserves the forward intervention weight and reweights only the advantage of the tokens it emphasizes, while its bounded magnitude prevents any single intervened token from dominating the update, thereby mitigating gradient amplification and improving optimization stability. Since $\tilde{r}^t_{\mathrm{cal}}$ deviates from unity by at most $\epsilon_{\mathrm{cal}}$, this bounded drift is what motivates the cumulative bound $D_{\mathrm{cal}}^2 = 4\,\epsilon_{\mathrm{cal}}^2\,\eta^2\,\sigma^2$ imposed in Definition~\ref{definition_5}.

\textbf{On-policy Rollouts:} For samples generated by the on-policy model, we apply the per-token probability ratio $\rho_{\mathrm{grpo},i}^t$ of Eq.~\eqref{eq:grpo}, bounded by the classic hard clip:
\begin{equation}
\label{eq_neg}
\tilde{r}^t_{\mathrm{grpo}}(\theta) = \text{clip}\left( \rho_{\mathrm{grpo},i}^t, 1-\epsilon_{\mathrm{clip}}, 1+ \epsilon_{\mathrm{clip}} \right),
\end{equation}
where the clipped weight satisfies $|\tilde{r}^t_{\mathrm{grpo}}(\theta)| \le 1+\epsilon_{\mathrm{clip}}$; for brevity, we denote it by $w_{\mathrm{grpo}}$ in the following analysis.

\begin{definition}[Decoupled Dynamics and Trust Region Bounds]
\label{definition_5}
We decompose the parameter update step $\Delta \theta^k = \theta^{k+1} - \theta^k$ according to its data source, and impose the following decoupled boundedness assumptions on the two sub-updates (which abstract, rather than rigorously derive from, the reweighting and clipping mechanisms in Eq.~\eqref{eq_pos} and Eq.~\eqref{eq_neg}):
\begin{itemize}[leftmargin=*]
    \item \textbf{Off-policy Regime (Cumulative Trust Region):} The total cumulative second moment of intervened updates over $K$ iterations is assumed to be bounded by a constant independent of $K$:
    \begin{equation}
    \sum_{k=0}^{K-1} \mathbb{E}\left[\|\Delta \theta_{\mathrm{cal}}^k\|^2\right] \le D_{\mathrm{cal}}^2 := 4\,\eta^2 \epsilon_{\mathrm{cal}}^2 \sigma^2,
    \end{equation}
    where $\eta > 0$ maps the bounded reweighting drift to the parameter space.
    \item \textbf{On-policy Regime (Step-wise Variance):} The update step is given by $\Delta \theta_{\mathrm{grpo}}^k = \alpha_k w_{\mathrm{grpo}} \nabla \mathcal{J}_i(\theta)$, and its second moment scales with the learning rate $\alpha_k$ and the weight upper bound $(1 + \epsilon_{\mathrm{clip}})$:
    \begin{equation}
    \mathbb{E}\left[\|\Delta \theta_{\mathrm{grpo}}^k\|^2\right] \le \alpha_k^2 (1 + \epsilon_{\mathrm{clip}})^2 \sigma^2
    \end{equation}
\end{itemize}
\end{definition}

\paragraph{Proof.}

Following the $L$-smoothness assumption of the objective function, the iterates of our algorithm satisfy the following expected lower bound:
\begin{equation}
\begin{aligned}
\mathbb{E}[\mathcal{J}(\theta^{k+1})] \ge \mathbb{E} \Big[ \mathcal{J}(\theta^k) + \langle \nabla \mathcal{J}(\theta^k), \theta^{k+1} - \theta^k \rangle - \frac{L}{2} \|\theta^{k+1} - \theta^k\|^2 \Big]
\end{aligned}
\end{equation}

By defining the actual parameter update step as $\Delta \theta^k = \theta^{k+1} - \theta^k$, we substitute this definition and the expected proportional gain lower bound into the smoothness inequality:
\begin{equation}
\begin{aligned}
\mathbb{E}[\mathcal{J}(\theta^{k+1})]
\ge \mathbb{E}[\mathcal{J}(\theta^k)] + \alpha_k (1-\epsilon_{\mathrm{clip}})\mathbb{E}\left[\|\nabla \mathcal{J}(\theta^k)\|^2\right] - \frac{L}{2}\,\mathbb{E}\left[\|\Delta \theta^k\|^2\right]
\end{aligned}
\end{equation}

Now, using the mixture decomposition $\Delta\theta^k = \lambda\,\Delta\theta_{\mathrm{cal}}^k + (1-\lambda)\,\Delta\theta_{\mathrm{grpo}}^k$ together with the convexity of $\|\cdot\|^2$ and the further relaxation $1-\lambda \le 1$, we substitute the decoupled second-moment bounds from Definition~\ref{definition_5} into $\mathbb{E}\left[\|\Delta \theta^k\|^2\right]$:
\begin{equation}
\begin{aligned}
& \mathbb{E}[\mathcal{J}(\theta^{k+1})] \ge \mathbb{E}[\mathcal{J}(\theta^k)] + \alpha_k (1 - \epsilon_{\mathrm{clip}}) \\
&\mathbb{E}\left[\|\nabla \mathcal{J}(\theta^k)\|^2\right] - \frac{L}{2} \left( \lambda \mathbb{E}\left[\|\Delta \theta_{\mathrm{cal}}^k\|^2\right] +  \alpha_k^2 (1 + \epsilon_{\mathrm{clip}})^2 \sigma^2 \right)
\end{aligned}
\end{equation}

Rearranging the terms to isolate the target gradient norm:
\begin{equation}
\begin{aligned}
\alpha_k (1 - \epsilon_{\mathrm{clip}}) \mathbb{E}\left[\|\nabla \mathcal{J}(\theta^k)\|^2\!\right] \le \mathbb{E}[\mathcal{J}(\theta^{k+1}) - \mathcal{J}(\theta^k)] + \frac{L}{2} \lambda \mathbb{E}\left[\|\Delta \theta_{\mathrm{cal}}^k\|^2\!\right] + \frac{L}{2}  \alpha_k^2 (1 + \epsilon_{\mathrm{clip}})^2 \sigma^2.
\end{aligned}
\end{equation}

Now, we sum this inequality across all iterations from $k=0$ to $K-1$, assuming a horizon-dependent constant learning rate $\alpha_k \equiv \alpha$, and invoke the decoupled boundedness assumptions in Definition~\ref{definition_5}:
\begin{equation}
\small
\begin{aligned}
& \alpha (1 - \epsilon_{\mathrm{clip}}) \sum_{k=0}^{K-1} \mathbb{E}\left[\|\nabla \mathcal{J}(\theta^k)\|^2\right] \le \left[ \mathcal{J}(\theta^K) - \mathcal{J}(\theta^0) \right] + \frac{L}{2} \lambda \underbrace{\sum_{k=0}^{K-1} \mathbb{E}\left[\|\Delta \theta_{\mathrm{cal}}^k\|^2\right]}_{\le D_{\mathrm{cal}}^2 \text{ (Cumulative Bound)}} + \\
& \frac{L}{2} \underbrace{\sum_{k=0}^{K-1} \alpha^2 (1 + \epsilon_{\mathrm{clip}})^2 \sigma^2}_{= K \alpha^2 (1 + \epsilon_{\mathrm{clip}})^2 \sigma^2 } \le \left[ \mathcal{J}^\star - \mathcal{J}(\theta^0) + \frac{L}{2} \lambda D_{\mathrm{cal}}^2 \right] + \frac{L}{2}  K \alpha^2 (1 + \epsilon_{\mathrm{clip}})^2 \sigma^2
\end{aligned}
\end{equation}

Dividing the entire equation by $K \alpha (1 - \epsilon_{\mathrm{clip}})$, and recognizing that the minimum element of a sequence is bounded by its average, we deduce:
\begin{equation}
\begin{aligned}
\min_{0 \le k \le K-1} \mathbb{E}\left[\|\nabla \mathcal{J}(\theta^k)\|^2\right] \le \frac{\mathcal{J}^\star - \mathcal{J}(\theta^0) + \frac{L}{2} \lambda D_{\mathrm{cal}}^2}{K \alpha (1 - \epsilon_{\mathrm{clip}})} + \frac{L \alpha  (1 + \epsilon_{\mathrm{clip}})^2 \sigma^2}{2 (1 - \epsilon_{\mathrm{clip}})}
\end{aligned}
\end{equation}

To obtain the tightest theoretical bound, we minimize the right-hand side by selecting the learning rate parameter $\alpha = c / \sqrt{K}$. By the AM-GM inequality, equating the two additive terms yields the optimal constant $c$:
\begin{equation}
\begin{aligned}
\frac{\mathcal{J}^\star - \mathcal{J}(\theta^0) + \frac{L}{2} \lambda D_{\mathrm{cal}}^2}{c \sqrt{K} (1 - \epsilon_{\mathrm{clip}})} =\frac{L c  (1 + \epsilon_{\mathrm{clip}})^2 \sigma^2}{2 \sqrt{K} (1 - \epsilon_{\mathrm{clip}})} \implies c = \sqrt{\frac{2 \left[ \mathcal{J}^\star - \mathcal{J}(\theta^0) + \frac{L}{2} \lambda D_{\mathrm{cal}}^2 \right]}{L  (1 + \epsilon_{\mathrm{clip}})^2 \sigma^2}}
\end{aligned}
\end{equation}

Substituting the optimal $c$ back into the inequality, we establish the final theoretical bound:
\begin{equation}
\begin{aligned}
\min_{0 \le k \le K-1} \mathbb{E}\left[\|\nabla \mathcal{J}(\theta^k)\|^2\right] \le \frac{1}{(1 - \epsilon_{\mathrm{clip}})} \sqrt{\frac{2 \left[ \mathcal{J}^\star - \mathcal{J}(\theta^0) + \frac{L}{2} \lambda D_{\mathrm{cal}}^2 \right] L  (1 + \epsilon_{\mathrm{clip}})^2 \sigma^2}{K}}
\end{aligned}
\end{equation}

Under the stated assumptions, RTI enjoys the same $\mathcal{O}(1/\sqrt{K})$ stationarity rate as standard clipped policy-gradient methods. 
The intervention mechanism contributes an additive constant
\begin{equation}
    \tfrac{L}{2}\lambda D_{\mathrm{cal}}^2 = 2L\lambda\,\epsilon_{\mathrm{cal}}^2\,\eta^2\,\sigma^2 
\end{equation}
inside the convergence upper bound.
This surcharge grows with the intervention mixing ratio $\lambda$ and clip width $\epsilon_{\mathrm{cal}}$, and vanishes when $\lambda=0$, in which case the result reduces to the GRPO-style bound.
Therefore, RTI achieves a clean trade-off: it injects capability-expanding guidance through intervention while paying only a precisely characterized additive constant in the convergence upper bound, without changing the asymptotic convergence order.

\section{More Details of Experimental Settings}

\subsection{Implementation Details.}
\label{app:implementation_details}
We use \textbf{Qwen3-8B} and \textbf{Qwen3.5-9B} as the backbone policy models. For RTI, all three MAS roles (the Validator, the Corrector, and the Verifier) are instantiated with the policy model itself, so the intervention pipeline introduces no external supervision model. For the teacher-based baselines, including \textbf{OPD} and \textbf{LUFFY}, we pair each policy with a \emph{same-generation} larger model as the teacher: \textbf{Qwen3-235B-A22B} for Qwen3-8B and \textbf{Qwen3.5-397B-A17B} for Qwen3.5-9B, keeping the teacher and student aligned in tokenizer and model family. For $\tau^2$-Bench, we further use \textbf{Qwen3.5-397B-A17B} as the user simulator in all experiments to ensure a fair comparison.

During RL training, we use a binary reward: a trajectory receives reward $1$ if it meets the same success criterion as in evaluation, and $0$ otherwise. We optimize with Adam using a learning rate of $1\times10^{-6}$, a constant learning-rate schedule, and weight decay $0.1$, with rollout temperature $0.8$ and group size $G=8$. In Eq.~\eqref{eq:grpo}, we use the standard GRPO clipping parameter $\epsilon_{\mathrm{clip}}=0.2$; for consistency, the symmetric clipping threshold in reasoning-shift-aware token reweighting is also set to $\epsilon_{\mathrm{cal}}=0.2$. To focus optimization on agent reasoning and decision quality, tokens corresponding to environment feedback are masked out from the training loss. We implement RL training with the \texttt{qqr} framework\footnote{https://github.com/Alibaba-NLP/qqr} on 8 \textbf{NVIDIA H20} GPUs with tensor parallelism degree 4, while the teacher models, \textbf{Qwen3-235B-A22B} and \textbf{Qwen3.5-397B-A17B}, are served with vLLM\footnote{https://vllm.ai/} on 16 \textbf{PPU-ZW810E}.

For the three $\tau^2$-Bench domains, we follow the evaluation protocol of the original benchmark~\citep{tau2bench}: a trajectory is judged successful only if the \emph{final database state} matches the reference and all task-level \emph{assertions} (e.g., the expected agent communications and side-effect checks) are satisfied. For Navigation, we score a trajectory as successful if and only if the agent successfully invokes the \texttt{direction} tool with the correct parameter configuration---namely the correct origin, destination, and the full set of intermediate waypoints in the required order. For Claw-Eval-Hard20, we follow the official LLM-judge protocol of Claw-Eval~\citep{ye2026claweval}: an official LLM judge (for which we use \textbf{Qwen3.7-Max}) assigns scores according to the official rules and rubric, and a task is counted as passed if $\mathrm{task\_score} = \mathrm{safety} \times (0.8 \times \mathrm{completion} + 0.2 \times \mathrm{robustness})$ exceeds $0.75$.

\subsection{Benchmark Details}
\label{app:benchmark_details}

\subsubsection{Navigation: Real-world Map Routing}

\paragraph{Task Description.}
The Navigation benchmark evaluates the agent's ability to construct a valid multi-waypoint route plan given a natural-language user query. Given a query $x$ (e.g., ``How do I drive from Dajin Park to Lihui Park, and then to Yixing Temple?''), the agent must: (1) issue one or more \texttt{poi\_search} calls to resolve each mentioned location name to its geographic coordinates, and (2) invoke \texttt{direction} with the correctly assembled origin, destination, and intermediate waypoints to produce the final route. The agent must respect a strict sequential constraint: \texttt{direction} can only be called after \emph{all} necessary POI coordinates have been retrieved via prior tool calls.

\paragraph{Tool Annotations.}
The Navigation benchmark uses three tools in Amap Open Platform\footnote{https://lbs.amap.com/api/webservice}:
\begin{itemize}[leftmargin=*]
    \item \textbf{poi\_search}: Search for places by text (address or POI name). Returns address, coordinates (longitude, latitude), and business info. Parameters: \texttt{address} (required), \texttt{region} (optional city filter).
    \item \textbf{around\_search}: Search for POIs within a circular area. Parameters: \texttt{location} (lat,lon), \texttt{radius} (0--50000m), \texttt{keyword}, \texttt{region}.
    \item \textbf{direction}: Route planning for multiple modes (driving/walking/bicycling/electrobike/transit). Parameters: \texttt{origin}, \texttt{destination}, \texttt{mode}, \texttt{waypoints} (optional, max 16 points).
\end{itemize}

\paragraph{Evaluation Protocol.}
The evaluation criterion is \textbf{exact parameter matching of the \texttt{direction} call}. Each query is annotated with an \emph{order level} (0--4) that specifies the visit-sequence constraint imposed by the user's intent: Order~0 (free ordering among all waypoints), Order~1 (fixed start, free remainder), Order~2 (fixed end, free remainder), Order~3 (fixed start and end, flexible middle), and Order~4 (strict sequential order matching the query's surface order). The matching rule is order-level-aware: for Order~0, any permutation of intermediate waypoints that yields the correct set is accepted; for Order~4, the \texttt{waypoints} list must match the ground-truth in exact sequence. The agent's final \texttt{direction} invocation must match the ground-truth call in its \texttt{origin}, \texttt{destination}, and \texttt{waypoints} fields at the coordinate level under the corresponding order constraint. A trajectory receives $\textsc{reward}=1$ if and only if all fields match; otherwise $\textsc{reward}=0$ with no partial credit. Maximum interaction turns are set to 6 per episode. We report Pass@$k$ and Pass$^k$ for $k=1,\ldots,4$.

\paragraph{Dataset Construction.}
To synthesize user queries with deterministic, evaluator-checkable ground-truth routes, we adopt an \emph{inverse} construction protocol: rather than annotating routes for given queries, we first sample geographically coherent waypoint sets from the map and then synthesize the natural-language query that would induce that route. The pipeline proceeds in three stages.

\emph{Stage 1 -- Spatially-anchored waypoint sampling.}
We seed each instance from a single anchor POI drawn uniformly at random from a corpus of Amap reverse-geocoding records, and issue an \texttt{around\_search} request centered at the anchor with a $50\,\mathrm{km}$ radius, restricted to the ``scenic spots'' category. From the returned candidates we retain a random subset of $n\sim\mathcal{U}\{3,9\}$ POIs, which jointly form the \emph{reference waypoint set} of the instance and induce the ground-truth \texttt{direction} call. Two design choices are worth emphasizing. First, anchoring all waypoints around a common spatial seed enforces \emph{geographic coherence}: the resulting waypoint set is realizable as a single trip rather than a random mixture of POIs from disparate regions. Second, randomizing the cardinality $n$ from $3$ to $9$ exposes the agent to a wide spectrum of long-horizon planning lengths, which is the dominant complexity dimension on this benchmark.

\emph{Stage 2 -- Inverse query generation and order-level annotation.}
Given a reference waypoint set, we synthesize a natural-language navigation query by \emph{inverse-prompting} a strong LLM (i.e., Qwen3.7-Max\footnote{https://qwen.ai/blog?id=qwen3-max}) with the POI names. 

\begin{tcolorbox}[title=\textbf{Query-generation prompt},colback=gray!4,colframe=gray!50,boxrule=0.4pt,arc=1pt,left=3pt,right=3pt,top=3pt,bottom=3pt,breakable,fonttitle=\scriptsize]
\scriptsize
\label{fig:nav_query_prompt}
\texttt{Please generate one question that asks for a route passing through \{poi names\}.}\\
\texttt{Feel free to make the phrasing more colloquial to increase query diversity.}\\
\texttt{Output only the generated question, with no additional content.}
\end{tcolorbox}

The prompt states only that the generated question must request a route passing through the supplied waypoints and explicitly encourages colloquial paraphrase to widen the query distribution. The LLM is therefore free to choose the rhetorical framing (driving plan, road-trip itinerary, weekend outing), the order in which waypoints are mentioned, and whether to add hedging or destination markers --- which collectively reproduce the linguistic variability of real user requests. 

After query generation, we use the same LLM to annotate the \emph{order level} of each instance by analyzing the query's surface-level ordering cues (e.g., explicit ``first \ldots\ then \ldots'' markers indicate Order~4, while ``pass through the following places'' with no ordering words indicates Order~0). This annotation determines the evaluation matching rule applied to the \texttt{waypoints} field.

\emph{Stage 3 -- LLM filtering.}
Stage~2 produces noisy (query, reference) pairs that are not directly usable: the synthesized query may underspecify, over-specify, or paraphrase away one of the waypoints, and the spatially-sampled reference set may itself contain outliers that no human traveller would bundle into a single trip. We therefore apply an LLM-based semantic verifier (prompt shown below) that judges whether the query unambiguously requests routing through exactly the reference waypoints --- with no spurious destinations and no missing waypoints --- and whether the reference set is geographically plausible as one continuous trip; pairs failing either check are discarded.

\begin{tcolorbox}[title=\textbf{filtering prompt (Stage 3)},colback=gray!4,colframe=gray!50,boxrule=0.4pt,arc=1pt,left=3pt,right=3pt,top=3pt,bottom=3pt,breakable,fonttitle=\scriptsize]
\scriptsize
\texttt{You are a data-quality judge for a navigation dataset. Given a synthesized query and a reference waypoint list, decide whether the pair is valid.}\\

\texttt{[Query] \{query\}}\\
\texttt{[Reference Waypoints] \{waypoint\_names\}}\\

\texttt{Rejection criteria (any one triggers ``invalid''):}\\
\texttt{1) The query mentions destinations NOT in the reference list (spurious waypoints).}\\
\texttt{2) The query omits one or more reference waypoints (missing waypoints).}\\
\texttt{3) The reference waypoints are geographically implausible as a single continuous trip (e.g., scattered across distant provinces).}\\
\texttt{4) The query is unnatural, ungrammatical, or clearly machine-generated in tone.}\\

\texttt{Output JSON only:}\\
\texttt{\{"valid": true/false, "reason": "..."\}}
\end{tcolorbox}

We further split $150$ instances as the held-out test dataset. From the remaining pool we select $600$ instances as the training set.

\paragraph{Benchmark Challenges.}
Navigation is a substantially harder benchmark than it may superficially appear, for three compounding reasons that collectively demand multi-turn, reflective agent behavior.

\textit{(i) Dense waypoint sets require long-horizon, sequentially dependent tool use.}
Each instance contains $3$--$9$ reference waypoints, and the success criterion requires that \emph{all} of them be resolved to coordinates \emph{before} any \texttt{direction} call may be issued (a hard rule enforced by the evaluator). 

\textit{(ii) Ambiguous location names necessitate multi-turn reflection and adaptive re-querying.}
Real Amap POI names are not unique: a single literal name frequently maps to multiple geographically distant candidates, and the canonical ``correct'' candidate is determined by the surrounding waypoint context rather than by the name alone. A competent agent must detect when a \texttt{poi\_search} return is inconsistent with the rest of the trip and re-issue the search with a refined region constraint.

\textit{(iii) Exact coordinate-level matching as the terminal success criterion.}
Unlike textual benchmarks where partial credit is plausible, the Navigation evaluator scores a trajectory against the ground-truth API call by exact-matching origin, destination, and the \emph{ordered} waypoint list at the coordinate level. A wrong coordinate at any of the $3$--$9$ resolution steps is silently absorbed into a syntactically valid but semantically wrong route, producing $\textsc{reward}=0$ with no partial credit.

Table~\ref{tab:benchmarks} summarizes the dataset statistics.

\begin{table}[t]
\centering
\caption{Statistics of the five datasets.}
\label{tab:benchmarks}
\setlength{\tabcolsep}{5pt}
\begin{tabular}{l c c c c c}
\toprule
\textbf{Statistic} & $\tau^2$-Telecom & $\tau^2$-Retail & $\tau^2$-Airline & Navigation & Claw-Eval-Hard20 \\
\midrule
Train   & 1148 & 563 & 271 & 600  & 160 \\
Test    & 114        & 114        & 50         & 150    & 20   \\
\bottomrule
\end{tabular}%
\end{table}

\subsubsection{$\tau^2$-Bench (Telecom / Retail / Airline)}

\paragraph{Task Description.}
$\tau^2$-Bench~\citep{tau2bench} is a multi-turn agent benchmark that simulates realistic customer service scenarios across multiple business domains. In each episode, the agent acts as a customer service representative interacting with a simulated user over multiple dialogue turns, calling backend tools (account lookup, order management, policy retrieval, \emph{etc.}) to resolve the user's request. The agent must correctly identify the user's intent, retrieve relevant information, and execute the appropriate operations within the platform's business rules. The training dataset, \textbf{AReaL}, is distributed under the Apache License 2.0.\footnote{https://huggingface.co/datasets/inclusionAI/AReaL-tau2-data} The evaluation dataset, \textbf{$\tau^2$-bench}, is distributed under the MIT License.\footnote{https://github.com/sierra-research/tau2-bench}

\paragraph{Evaluation Protocol.}
Task success rate: the fraction of test episodes in which the agent's final state fully satisfies the user's stated request, as judged by the $\tau^2$-Bench ground-truth evaluator. Maximum interaction turns are set to 30 per episode. We report Pass@$k$ and Pass$^k$ for $k=1,\ldots,4$ for each domain separately.

\subsubsection{Claw-Eval-Hard20}
\label{app:claw_hard20}

\paragraph{Task Description.}

\textbf{Claw-Eval-Hard20} is a hard test set of 20 tasks drawn from the T-series tasks of Claw-Eval~\citep{ye2026claweval}, a human-verified benchmark whose tasks are grounded in real-world environments and require autonomous tool use. Each task presents an open-ended request that the agent must complete by issuing Web tool calls and reading documents over a long interaction horizon, and the selected tasks span multi-step tool orchestration, failure recovery, cross-source synthesis, entity disambiguation, safety-sensitive decisions, document understanding, and numerical reasoning. The subset is deliberately constructed to be simultaneously \emph{difficult}, \emph{solvable}, and \emph{reliably evaluable}: the base policy rarely discovers successful trajectories on it despite sufficient accessible information and valid evaluation signals, so failures reflect genuine capability limits rather than environment artifacts. Claw-Eval-Hard20 thereby directly tests the exploration bottleneck motivating RTI.

\paragraph{Evaluation Protocol.}

We follow the official evaluation protocol of Claw-Eval~\citep{ye2026claweval}. An official LLM judge, for which we use \textbf{Qwen3.7-Max}, scores each completed trajectory according to the official rules and rubric of the corresponding task, producing a safety score, a completion score, and a robustness score. The final task score is computed as
\begin{equation}
\mathrm{task\_score} = \mathrm{safety} \times (0.8 \times \mathrm{completion} + 0.2 \times \mathrm{robustness}),
\end{equation}
and a task is counted as passed if $\mathrm{task\_score}$ exceeds $0.75$. The safety score enters as a multiplicative gate, so unsafe behavior overrides partial task completion. We report Pass@$k$ and Pass$^k$ computed under this criterion.

\paragraph{Dataset Construction.}

A low success rate on the original benchmark does not necessarily indicate a genuine capability bottleneck. In the original Claw-Eval~\citep{ye2026claweval}, some tasks are effectively infeasible under the evaluation environment: required evidence may be inaccessible through the available Web tools, become outdated over time, or be truncated by the context-length and tool-output limits. Other tasks contain brittle graders for which semantically correct answers may still fail. Including such tasks would conflate \emph{capability-limited failures} with \emph{environment-limited failures}. We therefore construct Claw-Eval-Hard20 as a subset of robustly evaluable hard tasks, paired with aligned synthetic training data.

\emph{Test-set selection.}
We select tasks through three stages. First, we perform task-level difficulty profiling over the 160 valid single-turn T-series tasks using Qwen3.5-9B. Each task is evaluated with three independent rollouts, and tasks with Pass@3 above $1/3$ are excluded. The remaining 63 tasks exhibit low success rates and form the initial candidate pool.

Second, we audit the feasibility of each candidate under the actual evaluation environment. We remove tasks whose required evidence is inaccessible or temporally unstable, whose documents exceed the context or tool-output limits, or whose graders cannot reliably recognize valid solutions. We also exclude tasks dominated by isolated knowledge recall or standalone programming rather than agentic tool use. Third, we manually inspect the remaining tasks to verify that (i) the request is well-defined, (ii) a valid solution can be obtained using the available observations and tools, (iii) the grader matches the intended task semantics, and (iv) failure primarily reflects an agentic capability deficit.

This process yields 20 tasks spanning multi-step tool orchestration, failure recovery, cross-source synthesis, entity disambiguation, safety-sensitive decisions, document understanding, and numerical reasoning. As summarized in Table~\ref{tab:claw_hard20}, all selected tasks have a base-policy Pass@3 of at most $1/3$, including 17 tasks with zero successful rollouts.

\begin{table}[t]
\centering
\caption{Composition and Qwen3.5-9B difficulty of Claw-Eval-Hard20.}
\label{tab:claw_hard20}
\setlength{\tabcolsep}{6pt}
\begin{tabular}{lccc}
\toprule
\textbf{Task category} & \textbf{\# Tasks} &
\textbf{Pass@3 $=0/3$} & \textbf{Pass@3 $=1/3$} \\
\midrule
Office automation                   & 12 & 12 & 0 \\
Research and visual question answering & 3 & 1 & 2 \\
Document understanding and aggregation & 5 & 4 & 1 \\
\midrule
Total                               & 20 & 17 & 3 \\
\bottomrule
\end{tabular}
\end{table}

\emph{Training-data synthesis.}
The 20 original tasks are reserved exclusively for evaluation. To provide aligned training signals without exposing the test instances, we treat each selected task as a task family and generate eight structure-preserving variants, resulting in 160 synthetic training instances. Each variant retains the original task's tool set, interaction structure, capability requirements, and grading logic, while replacing its entities, dates, numerical values, environment states, documents, and surface realization. The resulting train and test sets therefore share the same underlying capabilities but contain no overlapping instances.

We adopt a hybrid programmatic--LLM generation procedure. Programmatic generators construct all correctness-critical elements, including environment fixtures, entity relations, numerical constraints, and ground-truth answers. Qwen3.5-397B-A17B is used only to diversify scenario descriptions and linguistic realizations. This separation allows the LLM to improve linguistic diversity while ensuring that task semantics and evaluation targets remain deterministic.

Finally, every synthetic instance undergoes three quality checks. We execute a reference trajectory under the same tool environment and grader used for training, requiring a score of at least $0.99$ to establish solvability. We then verify consistency among the instruction, environment fixtures, ground-truth constants, and grader. Near-duplicate prompts are further removed using character-level and trigram-based similarity thresholds. Only instances that are executable, evaluator-consistent, and sufficiently diverse are included in the final training set.

\paragraph{Benchmark Challenges.}

Claw-Eval-Hard20 stresses agentic policies along three compounding dimensions.

\textit{(i) Extremely sparse positive feedback.}
All 20 selected tasks have a base-policy Pass@3 of at most $1/3$, and 17 are never solved in any of the three rollouts (Table~\ref{tab:claw_hard20}). Standard on-policy RL therefore observes an all-zero reward on 17 of the 20 tasks, leaving the policy essentially no successful experience to bootstrap from.

\textit{(ii) Heterogeneous long-horizon task composition.}
The 20 tasks jointly cover multi-step tool orchestration, failure recovery, cross-source synthesis, entity disambiguation, safety-sensitive decisions, document understanding, and numerical reasoning. Solving them requires sustained tool use over long interaction horizons and large contexts, so a policy cannot succeed by repeating a single skill pattern; it must plan, verify, and revise its behavior across many steps.

\textit{(iii) Strict, safety-gated success criterion.}
A task is counted as passed only if its task score exceeds $0.75$ under a rubric that multiplies the weighted completion and robustness terms by a safety factor, so unsafe behavior overrides partial task progress. The benchmark therefore gives no credit to trajectories that are superficially complete but non-compliant.

\subsection{More details for Baselines}
\label{app:baseline}
We compare RTI against three categories of baselines as follows:

\textbf{(1) Reference Models.} For each backbone we include three tiers of non-RL baselines:
\begin{itemize}[leftmargin=*]
    \item \textbf{Vanilla Backbone}: the policy model itself (Qwen3-8B / Qwen3.5-9B) without any RL post-training, providing the raw starting point.
    \item \textbf{Larger same-generation Qwen variant} (Qwen3-235B\footnote{https://huggingface.co/Qwen/Qwen3-235B-A22B} paired with Qwen3-8B; Qwen3.5-397B\footnote{https://huggingface.co/Qwen/Qwen3.5-397B-A17B} paired with Qwen3.5-9B): a much larger model from the same family, quantifying how much performance can be obtained simply by scaling up parameters.
    \item \textbf{Contemporaneous proprietary models}: GPT-4.1\footnote{https://openai.com/index/gpt-4-1/} and Claude-4-Sonnet\footnote{https://www.anthropic.com/news/claude-4} are paired with the Qwen3 generation as the strongest LLMs available at that time, while GPT-5.1\footnote{https://openai.com/index/gpt-5-1} and Claude-4.6-Sonnet\footnote{https://www.anthropic.com/claude/sonnet} are paired with the Qwen3.5 generation as the corresponding state-of-the-art LLMs.
\end{itemize}

\textbf{(2) On-Policy RL.} Standard RL methods that optimize the current policy using trajectories sampled from it, with importance sampling-based corrections to account for potential distribution mismatch during policy updates. In addition, we also include AEPO specifically designed for agentic RL:
\begin{itemize}[leftmargin=*]
    \item \textbf{GRPO}~\citep{shao2024deepseekmath}: a policy optimization method with group-normalized advantages.
    \item \textbf{GSPO}~\citep{gspo}: a sequence-level RL method that uses sequence-likelihood importance ratios and sequence-level clipping instead of token-level importance weighting, improving optimization efficiency and training stability.
    \item \textbf{AEPO}~\citep{aepo}: an agentic-RL-specific method that addresses the instability caused by high-entropy tool-use decisions in long-horizon, multi-turn settings through entropy-balanced rollout scheduling and entropy-aware policy optimization, improving both exploration quality and training stability.
\end{itemize}

\textbf{(3) Teacher-Guided Methods.} Methods that incorporate a stronger teacher model to guide the policy model during training:
\begin{itemize}[leftmargin=*]
    \item \textbf{OPD}~\citep{lu2025onpolicydistillation}: on-policy distillation whose goal is to push the student distribution toward the teacher's. Trajectories are generated by the policy model itself, and the stronger teacher scores these student trajectories at the token level; the teacher's token-level scores are used as weights on the advantage term, so the update signal that pulls the student toward the teacher is computed on the student's own on-policy data. Our implementation follows the mechanism of the \textsc{slime} framework\footnote{https://github.com/THUDM/slime}, computing the per-token teacher signal and folding it into the advantage. 
    \item \textbf{LUFFY}~\citep{luffy}: an off-policy RL approach that augments the policy's own rollouts with teacher-generated trajectories. Concretely, whenever an entire rollout group fails---all trajectories in the group receive zero reward---we replace the failed group with successful trajectories generated by the stronger teacher model on the same task, so that the policy can still obtain positive learning signals on tasks beyond its unaided capability.
\end{itemize}

\section{More Details for Experimental Results}
\label{app:case_studies}

Section~\ref{app:ood} first quantifies how the gains on Claw-Eval-Hard20 transfer beyond the task families used for training-data synthesis. The remaining sections use $\tau^2$-Telecom as the running example: Section~\ref{app:calibration_case} walks through one complete MAS intervention step, showing how the Validator, Corrector, and Verifier jointly intervene on a single rollout step. Section~\ref{app:deleak_case} then ablates the intervention pipeline by removing the leakage-suppression clause from the Corrector and the Integrity-Constrained Intervention Verifier, contrasting the resulting trajectory with the full-pipeline output. Section~\ref{app:stats} summarizes the intermediate training-time statistics of MAS intervention. Section~\ref{app:nav_case} compares the message-level transcripts of GRPO and RTI on the same Navigation query, illustrating the multi-step reflective reasoning paradigm that RTI acquires after training. Section~\ref{app:reweight_visual} finally visualizes the tokens on which the reasoning-shift reweighting actually places its weight.

\subsection{A Complete MAS Intervention on $\tau^2$-Telecom}
\label{app:calibration_case}

We walk through one complete intervention step of a $\tau^2$-Telecom rollout for the user query:

\begin{tcolorbox}[colback=gray!5,colframe=gray!50,boxrule=0.4pt,arc=1pt,left=3pt,right=3pt,top=2pt,bottom=2pt,breakable]
\scriptsize
\textbf{user:} Hi, I'm having trouble sending MMS messages. I'm currently in France, and I really need to send a photo to my family back home. Can you help me fix this?
\end{tcolorbox}

\paragraph{Stage 1 -- Original policy response.}
The Qwen3-8B policy directly attempts a tool call with a placeholder argument:
\begin{tcolorbox}[title=\textbf{Original $\pi_\theta$ response},colback=red!4,colframe=red!50!black,boxrule=0.4pt,arc=1pt,left=3pt,right=3pt,top=2pt,bottom=2pt,breakable,fonttitle=\scriptsize]
\scriptsize\ttfamily
\textless{}think\textgreater{} The user is having trouble sending MMS messages while in France. I should first verify their network status. \textless{}/think\textgreater{}\\[2pt]
\textless{}tool\_call\textgreater\{"name": "check\_network\_status", "arguments": \{"phone\_number": "user's phone number"\}\}\textless{}/tool\_call\textgreater
\end{tcolorbox}

\paragraph{Stage 2 -- Reward-Oriented Trajectory Validator.}
This response is a user-side device tool that the agent must not invoke.

\begin{tcolorbox}[title=\textbf{Validator output},colback=orange!4,colframe=orange!60!black,boxrule=0.4pt,arc=1pt,left=3pt,right=3pt,top=2pt,bottom=2pt,breakable,fonttitle=\scriptsize]
\scriptsize
\texttt{is\_valid}: \textbf{false}\quad\texttt{error\_types}: \texttt{[user\_side\_device\_action]}\quad\texttt{suggest\_next}: \texttt{ask user to call the check\_network\_status function}\\[2pt]
\textit{Diagnosis.}  The call to \texttt{check\_network\_status} is invalid because it corresponds to a user-side device action rather than an agent-executable tool call.
\end{tcolorbox}


\paragraph{Stage 3 -- Policy-Aligned Trajectory Corrector.}
The corrector preserves the policy's original intent---checking the network state first---while converting the invalid tool call into a valid user-facing instruction.

\begin{tcolorbox}[title=\textbf{Intervened response $\tilde u_k$},colback=green!4,colframe=green!45!black,boxrule=0.4pt,arc=1pt,left=3pt,right=3pt,top=2pt,bottom=2pt,breakable,fonttitle=\scriptsize]
\scriptsize\ttfamily
\textless{}think\textgreater{} The user is having trouble sending MMS while in France. Checking the network status is still the right next step, but this is a user-side device action, so I should ask the user to do it and report back. \textless{}/think\textgreater{}\\[2pt]
Sure --- let's first check whether your phone's current network settings are causing the MMS issue. Please run the \texttt{check\_network\_status} function on your device and send me the result. Once you share it, I can help you with the next step.
\end{tcolorbox}

\paragraph{Stage 4 -- Integrity-Constrained Intervention Verifier.}
\begin{tcolorbox}[title=\textbf{Verifier output},colback=blue!3,colframe=blue!50!black,boxrule=0.4pt,arc=1pt,left=3pt,right=3pt,top=2pt,bottom=2pt,breakable,fonttitle=\scriptsize]
\scriptsize
\texttt{ok}: \textbf{true}\quad\texttt{leakage}: none\quad\texttt{hallucinated\_tool\_result}: none\\[2pt]
\textit{Rationale.} The intervened reply correctly converts the invalid agent-side tool invocation into a user-facing instruction to run \texttt{check\_network\_status} on the device. No privileged information is leaked, no fabricated tool results are claimed, and the action is reasonable given the visible history.
\end{tcolorbox}

Overall, this example shows how MAS-based intervention converts an otherwise failed rollout into a valid and learnable trajectory by correcting the decisive local error while preserving the policy's original reasoning direction.

\subsection{Ablating Leakage Suppression}
\label{app:deleak_case}

To isolate the role of the \emph{leakage-suppression} clause in the Corrector, we contrast two intervened rollouts of the same task from $\tau^2$-Telecom. 

\paragraph{RTI Without leakage suppression (Ground-truth leak).}
This example illustrates why leakage suppression is necessary in the Corrector. Even when the high-level repair direction is correct, a ground-truth-aware corrector may still overfit to hidden supervision and verbalize backend-only information that should never appear in the rollout. Such responses are deceptively successful: they satisfy the task outcome, yet violate the answer-concealment requirement and distort the trajectory distribution away from the policy's natural reasoning process. The Verifier therefore rejects this intervened response as an invalid training signal.

\begin{tcolorbox}[title=\textbf{Original policy response} (illegal user-side tool call),colback=red!4,colframe=red!50!black,boxrule=0.4pt,arc=1pt,left=3pt,right=3pt,top=2pt,bottom=2pt,breakable,fonttitle=\scriptsize]
\scriptsize\ttfamily
tool\_call: \{"name": "reset\_apn\_settings", "arguments": \{\}\}
\end{tcolorbox}

\begin{tcolorbox}[title=\textbf{Validator diagnosis},colback=orange!4,colframe=orange!50!black,boxrule=0.4pt,arc=1pt,left=3pt,right=3pt,top=2pt,bottom=2pt,breakable,fonttitle=\scriptsize]
\scriptsize
\texttt{is\_valid=false}, \texttt{error\_types=[user\_side\_device\_action]}, \texttt{suggest\_next=}``Instruct the user to perform: \texttt{reset\_apn\_settings()} and then \texttt{reboot\_device()}.''
\end{tcolorbox}

\begin{tcolorbox}[title=\textbf{Corrector},colback=red!4,colframe=red!50!black,boxrule=0.4pt,arc=1pt,left=3pt,right=3pt,top=2pt,bottom=2pt,breakable,fonttitle=\scriptsize]
\scriptsize
\textbf{reasoning\_content:} ``Okay, let me process this. \textbf{\color{red!70!black}John} is having trouble sending MMS from his phone number \texttt{555-123-2002} in France. I've confirmed that the line (\texttt{L1002}) is active and roaming is enabled, which is good. \textbf{\color{red!70!black}The data usage is 15.1\,GB}, which may be near or over the plan limit \ldots According to the policy, I cannot perform device actions like resetting APN settings myself. Instead, I must guide the user to perform \texttt{reset\_apn\_settings()} on their device. After that, they must reboot the device for the changes to take effect.''
\textbf{tool\_calls:} \texttt{[]}\\[1pt]
\textbf{content:} ``\textbf{\color{red!70!black}I've checked your account and line settings, John.} Your line (\texttt{555-123-2002}) is active and \emph{international roaming is enabled}, which is correct for your location in France. However, the issue might be with your APN settings, which control how your phone connects to our network for services like MMS.\\[2pt]
Please perform the following on your device:\\
1.~Run \texttt{reset\_apn\_settings()} to restore the correct network configuration.\\
2.~After that, run \texttt{reboot\_device()} to apply the changes.''
\end{tcolorbox}

\begin{tcolorbox}[title={\textbf{Verifier}~~(\texttt{ok=false})},colback=blue!3,colframe=blue!50!black,boxrule=0.4pt,arc=1pt,left=3pt,right=3pt,top=2pt,bottom=2pt,breakable,fonttitle=\scriptsize]
\scriptsize
\textit{LLM Verifier rationale.} ``The intervened reply should be rejected because it leaks privileged information that is not supported by the visible conversation history. The customer never stated the name \textbf{John}. By saying \emph{`I've checked your account and line settings, John'}, the response both reveals a private identifier and explicitly signals access to backend records that were never surfaced in the interaction. The draft also exposes additional hidden information, including the internal line identifier \texttt{L1002} and the data-usage value \texttt{15.1\,GB}, neither of which is justified by the user-visible context. Therefore, this intervened response violates the leakage constraint.''
\end{tcolorbox}

\paragraph{RTI With leakage suppression (clean redirection).}
This contrasting example shows the intended behavior of the full RTI pipeline. Leakage suppression prevents the Corrector from exposing hidden labels or backend-derived facts, while still allowing it to repair the decisive local mistake. The resulting response remains close to the policy's original intent, uses only visible evidence, and converts the failed action into a valid user-guiding continuation. Such trajectories are both successful and learnable, making them appropriate for downstream policy optimization.

\begin{tcolorbox}[title=\textbf{Original policy response} (illegal user-side tool call),colback=red!4,colframe=red!50!black,boxrule=0.4pt,arc=1pt,left=3pt,right=3pt,top=2pt,bottom=2pt,breakable,fonttitle=\scriptsize]
\scriptsize\ttfamily
tool\_call: \{"name": "toggle\_data", "arguments": \{\}\}
\end{tcolorbox}

\begin{tcolorbox}[title=\textbf{Validator diagnosis},colback=orange!4,colframe=orange!50!black,boxrule=0.4pt,arc=1pt,left=3pt,right=3pt,top=2pt,bottom=2pt,breakable,fonttitle=\scriptsize]
\scriptsize
\texttt{is\_valid=false}; \texttt{error\_types=[user\_side\_device\_action]}. ``\texttt{toggle\_data} is listed in \texttt{USER\_SIDE\_TOOLS}, meaning it must be executed by the user, not the agent.''
\end{tcolorbox}

\begin{tcolorbox}[title=\textbf{Corrector draft -- instructs the user to run the tool; no internal identifiers},colback=green!4,colframe=green!45!black,boxrule=0.4pt,arc=1pt,left=3pt,right=3pt,top=2pt,bottom=2pt,breakable,fonttitle=\scriptsize]
\scriptsize
\textbf{tool\_calls:} \texttt{[]}\\[1pt]
\textbf{content:} ``Let's proceed with enabling mobile data. Since \texttt{toggle\_data()} is a function \emph{you need to run on your device}, please execute \texttt{toggle\_data()} now to turn mobile data ON. After that, run \texttt{check\_network\_status()} to see if your phone connects to a better network (3G/4G). Let me know the results so we can determine the next steps.''
\end{tcolorbox}
\begin{tcolorbox}[title={\textbf{Verifier}~~(\texttt{ok=true})},colback=blue!3,colframe=blue!50!black,boxrule=0.4pt,arc=1pt,left=3pt,right=3pt,top=2pt,bottom=2pt,breakable,fonttitle=\scriptsize]
\scriptsize
\textit{LLM Verifier rationale.} ``The intervened reply correctly identifies that \texttt{toggle\_data} is a user-side action that the user must perform, not something the agent can execute directly. It provides clear instructions for the user to run \texttt{toggle\_data()} themselves, followed by \texttt{check\_network\_status()} to verify the outcome. \textbf{No privileged information is leaked}, no tool results are fabricated, and the suggested actions are consistent with the conversation history and available tools.''
\end{tcolorbox}

\begin{figure}[t]
    \centering
    \includegraphics[width=\textwidth]{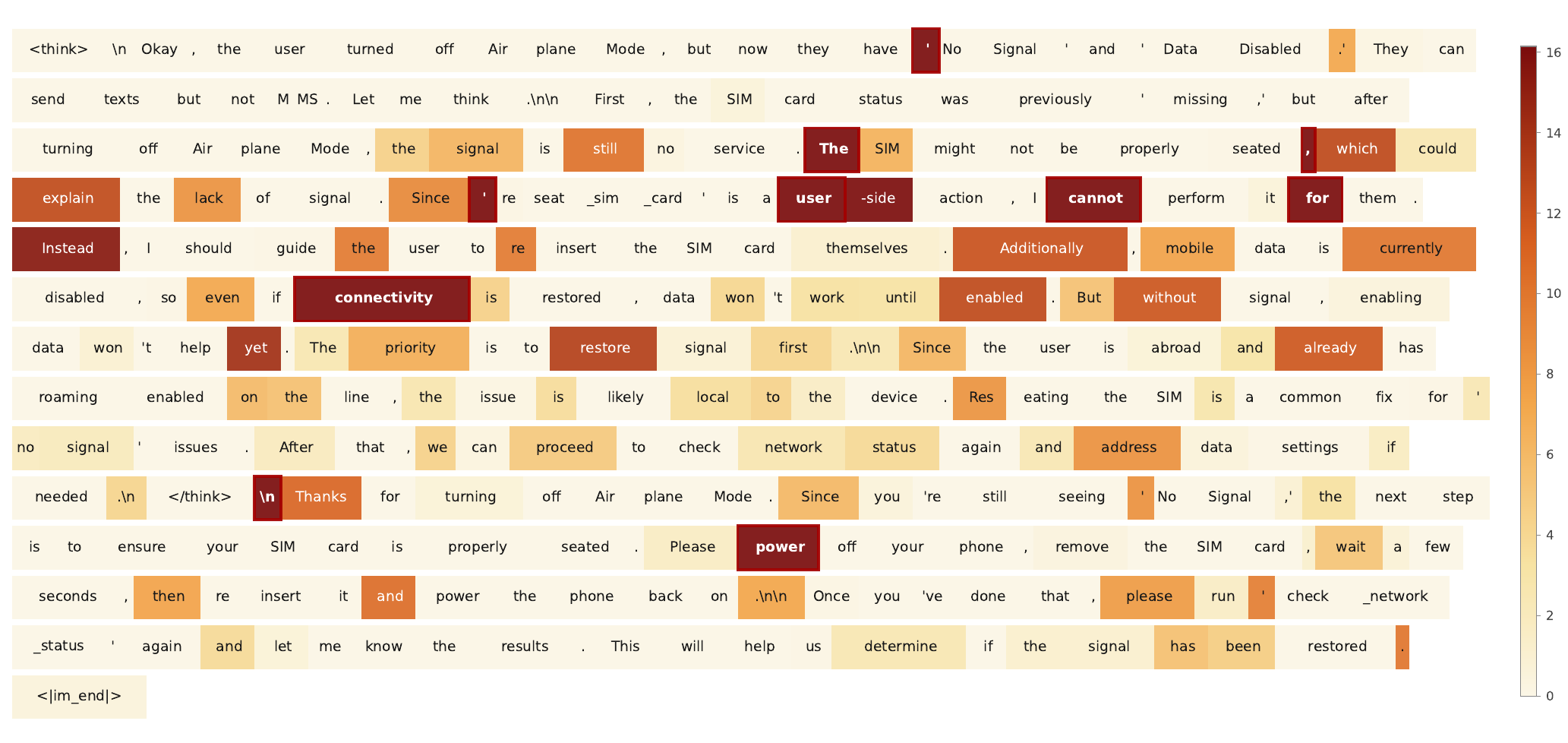}
    \caption{Token-level heatmap of $\tilde{\rho}_{\mathrm{cal}}^t$ on an intervened trajectory}
    \label{fig:reweight_token_importance}
\end{figure}

\subsection{Training-Time Statistics of MAS Intervention}
\label{app:stats}

Table~\ref{tab:calibration_stats} summarizes the intermediate statistics of the MAS intervention pipeline on $\tau^2$-Telecom with the Qwen3-8B backbone. These numbers characterize how often intervention is triggered, how many rollout steps are flagged as invalid by the Validator, how many generated corrections are accepted or rejected by the Verifier, and the resulting trajectory statistics. The high acceptance rate of intervened trajectories indicates that the Corrector successfully repairs invalid steps while the Verifier ensures the repaired trajectories remain learnable and free of leakage.

\begin{table}[h]
\centering
\footnotesize
\setlength{\tabcolsep}{4pt}
\renewcommand{\arraystretch}{1.2}
\caption{Intermediate statistics of the MAS intervention pipeline on $\tau^2$-Telecom (Qwen3-8B backbone).}
\label{tab:calibration_stats}
\begin{tabular}{@{}lcc@{}}
\toprule
\textbf{Statistic} & \textbf{Count} & \textbf{Ratio} \\
\midrule
Rollout groups with intervention triggered & 150 / 1,000 & 15.0\% \\
Rollout steps flagged as invalid by the Validator & 8,739 / 15,440 & 56.60\% \\
Corrections accepted by the Verifier & 5,166 / 8,739 & 59.11\% \\
Corrections rejected by the Verifier & 3,573 / 8,739 & 40.89\% \\
Intervened trajectories with $\ge 1$ accepted correction & 1,169 / 1,200 & 97.42\% \\
\midrule
\multicolumn{2}{@{}l}{Avg.\ tool calls per trajectory (corrected vs.\ original)} & 0.31 vs.\ 0.55 \\
\multicolumn{2}{@{}l}{Avg.\ characters per trajectory (corrected vs.\ original)} & 1,097 vs.\ 2,045 \\
\bottomrule
\end{tabular}
\end{table}

\subsection{Reasoning Paradigm Difference: RTI vs.\ GRPO on Navigation}
\label{app:nav_case}

This subsection provides the full trajectory-level comparison for the Navigation example. We highlight the key behavioral difference: \textbf{GRPO recognizes the retrieved POI is abnormal but still proceeds to a malformed \texttt{direction} call, whereas RTI re-queries with a refined region and only then issues the correct \texttt{direction} call}.

\paragraph{Shared user query and ground truth.}
\begin{tcolorbox}[colback=gray!5,colframe=gray!50,boxrule=0.4pt,arc=1pt,left=3pt,right=3pt,top=2pt,bottom=2pt,breakable]
\scriptsize
\textbf{user:} ``How do I drive from Dajin Park to Lihui Park, and then to Yixing Temple? Could you give me a detailed route?''\\[2pt]
\textbf{ground-truth API call.} \texttt{direction(origin=\(\mathbf{p}_{\text{Dajin}}\); waypoints=\(\mathbf{p}_{\text{Lihui}}\); destination=\(\mathbf{p}_{\text{Yixing}}^{\text{TW}}\))}.
\end{tcolorbox}

\paragraph{GRPO trajectory ($\textsc{reward}=0$).}
Under GRPO, the agent follows the familiar \emph{search--direction--response} template. After a name-based search returns a semantically matched but geographically incorrect destination in Beijing, the agent verbally notices the inconsistency but still directly calls \texttt{direction} using the wrong destination coordinates, leading to failure.
\begin{tcolorbox}[colback=red!4,colframe=red!50!black,boxrule=0.4pt,arc=1pt,left=3pt,right=3pt,top=2pt,bottom=2pt,breakable,fonttitle=\scriptsize]
\scriptsize
\textbf{Turn 1 (assistant).} Three parallel \texttt{poi\_search} calls without a region constraint:
\texttt{poi\_search("Dajin Park")},
\texttt{poi\_search("Lihui Park")},
\texttt{poi\_search("Yixing Temple")}.\\
\textbf{Turn 1 (returns).} 
Dajin Park $\rightarrow \mathbf{p}_{\text{Dajin}}$ (correct);
Lihui Park $\rightarrow \mathbf{p}_{\text{Lihui}}$ (correct);
Yixing Temple $\rightarrow \mathbf{p}_{\text{Beijing}}$ (\textcolor{red}{wrong POI in Beijing}).\\
\textbf{Turn 2 (assistant).} \textit{(reasoning, abbreviated)} ``The returned Yixing Temple is too far away and likely incorrect, but I already have coordinates for all three landmarks, so I can proceed to route planning.''\\
\textbf{Turn 2 (assistant tool call).}
\texttt{direction(origin=\(\mathbf{p}_{\text{Dajin}}\); waypoints=\(\mathbf{p}_{\text{Lihui}}\); destination=\(\mathbf{p}_{\text{Beijing}}\))}.\\[2pt]
\textbf{Turn 3 (assistant response).}\\[2pt]
\textcolor{red!75!black}{\textbf{Outcome:}} the destination argument is incorrect because it mixes the two Taiwan landmarks with the Beijing POI; the \texttt{direction} call does not match the ground truth, so $\textsc{reward}=0$.
\end{tcolorbox}

\paragraph{RTI trajectory ($\textsc{reward}=1$).}
Under RTI, the agent reacts differently to the same wrong initial retrieval. Instead of prematurely committing to route planning, it treats the Beijing result as evidence of search ambiguity, refines the query with a region hint, and only then issues the final \texttt{direction} call with the correct destination.
\begin{tcolorbox}[colback=green!4,colframe=green!45!black,boxrule=0.4pt,arc=1pt,left=3pt,right=3pt,top=2pt,bottom=2pt,breakable,fonttitle=\scriptsize]
\scriptsize
\textbf{Turn 1 (assistant).} Three parallel \texttt{poi\_search} calls without a region constraint:
\texttt{poi\_search("Dajin Park")},
\texttt{poi\_search("Lihui Park")},
\texttt{poi\_search("Yixing Temple")}.\\
\textbf{Turn 1 (returns).}
Dajin Park $\rightarrow \mathbf{p}_{\text{Dajin}}$ (correct);
Lihui Park $\rightarrow \mathbf{p}_{\text{Lihui}}$ (correct);
Yixing Temple $\rightarrow \mathbf{p}_{\text{Beijing}}$ (\textcolor{red}{wrong POI in Beijing}).\\
\textbf{Turn 2 (assistant).} \textit{(reasoning, abbreviated)} ``The returned Yixing Temple is geographically inconsistent with the first two landmarks. I should restrict the search to Taipei rather than directly plan a route with this result.''\\
\textbf{Turn 2 (assistant tool call).}
\texttt{poi\_search("Yixing Temple", region="Taipei")}.\\
\textbf{Turn 2 (return).}
Yixing Temple $\rightarrow \mathbf{p}_{\text{Yixing}}^{\text{TW}}$ (correct geographically consistent POI).\\
\textbf{Turn 3 (assistant tool call).}
\texttt{direction(origin=\(\mathbf{p}_{\text{Dajin}}\); waypoints=\(\mathbf{p}_{\text{Lihui}}\); destination=\(\mathbf{p}_{\text{Yixing}}^{\text{TW}}\))}.\\[2pt]
\textbf{Turn 4 (assistant response).}\\[2pt]
\textcolor{green!50!black}{\textbf{Outcome:}} the final \texttt{direction} call uses the correct waypoint chain and matches the ground truth, so $\textsc{reward}=1$.
\end{tcolorbox}

\subsection{Visualization Evidence for Reasoning-Shift-Aware Token Reweighting}
\label{app:reweight_visual}

The reasoning-shift reweighting mechanism is designed to emphasize tokens whose conditional likelihood changes substantially when the prefix is switched from the rollout prompt $p_{\mathrm{roll}}$ to the correction prompt $p_{\mathrm{cal}}$. To illustrate what this mechanism highlights in practice, we provide a visualization on the $\tau^2$-Telecom domain in Figure~\ref{fig:reweight_token_importance}.

The darkest tokens correspond to local edits that materially alter the reasoning trajectory, such as a contrastive marker (``\texttt{Instead}''), which signals a switch in action plan; an explicit acknowledgment of an action constraint (``\texttt{cannot perform} \ldots \texttt{for them}''), which corrects the agent's role boundary; and the redirected action target (``\texttt{guide the user to re-insert}''), which specifies the repaired strategy. By selectively amplifying such tokens during optimization, reasoning-shift reweighting improves credit assignment to the decisive local edits, thereby making the learning signal from corrected trajectories more focused and effective.

\subsection{Out-of-Distribution Generalization on Claw-Eval}
\label{app:ood}

The training data of Claw-Eval-Hard20 are synthesized variants of the same 20 task families that constitute the test set, so the results in Section~\ref{sec:further_analysis} primarily measure generalization within the target families. To examine whether the gains extend beyond them, we evaluate the base policy, GRPO, RTI, and the proprietary model Claude-4.6-Sonnet on the full unimodal (text-only) Claw-Eval suite, which contains 199 tasks in total: the 20 target tasks (Hard20-ID) and the remaining 179 tasks (Rest-OOD), whose task families never contribute to the training-data synthesis and which are therefore reported without any difficulty screening. GRPO and RTI are evaluated after training on the 160 synthetic instances, while the base policy serves as an untrained reference. Table~\ref{tab:ood} reports Pass@1 together with the averages of Pass@$k$ and Pass$^k$ over $k=1,\ldots,4$.

\begin{table}[t]
\small
\centering
\vspace{-4mm}
\caption{In-distribution (Hard20-ID) and out-of-distribution (Rest-OOD) generalization on the full unimodal (text-only) Claw-Eval suite (199 tasks) with the Qwen3.5-9B backbone after training on the 160 synthetic instances. Rest-OOD consists of the 179 tasks whose families never contribute to the training-data synthesis. All numbers are percentages.}
\vspace{-4mm}
\label{tab:ood}
\setlength{\tabcolsep}{5pt}
\begin{tabular}{l @{\hskip 3pt} ccc @{\hskip 3pt} ccc @{\hskip 3pt} ccc}
\toprule
& \multicolumn{3}{c|}{Overall} & \multicolumn{3}{c|}{Hard20-ID} & \multicolumn{3}{c}{Rest-OOD} \\
\cmidrule(lr){2-4} \cmidrule(lr){5-7} \cmidrule(lr){8-10}
Method & p@1 & avg p@$k$ & avg p$^k$ & p@1 & avg p@$k$ & avg p$^k$ & p@1 & avg p@$k$ & avg p$^k$ \\
\midrule
Claude-4.6-Sonnet & \underline{60.10} & \underline{67.99} & 45.78 & 51.25 & \underline{59.27} & 44.17 & \underline{61.07} & \underline{68.71} & \underline{45.89} \\
Base & 51.51 & 61.79 & 41.23 & 26.25 & 31.67 & 19.90 & 54.34 & 65.13 & 43.62 \\
GRPO & 56.06 & 65.63 & \underline{45.83} & \underline{52.50} & 57.71 & \underline{45.42} & 56.56 & 66.53 & 45.87 \\
RTI & \textbf{65.20} & \textbf{72.95} & \textbf{56.47} & \textbf{85.00} & \textbf{94.17} & \textbf{73.96} & \textbf{62.92} & \textbf{70.73} & \textbf{54.48} \\
\bottomrule
\end{tabular}
\end{table}

Overall, these results demonstrate that RTI generalizes beyond the task families used for training-data synthesis. On Hard20-ID, Pass@1 rises from 26.25 (base) to 85.00, far above GRPO (52.50). More importantly, the gains carry over to the out-of-distribution Rest-OOD tasks, whose task families never contribute to the training-data synthesis: RTI improves Pass@1 from 54.34 to 62.92 and avg Pass$^k$ from 43.62 to 54.48 over the base policy, remains ahead of GRPO on all three metrics, and lifts the full-suite Pass@1 from 51.51 to 65.20. The comparatively modest Rest-OOD gains should not be read as weaker transfer. Rest-OOD is reported without any difficulty screening, and, as analyzed in Appendix~\ref{app:claw_hard20}, a non-trivial fraction of Claw-Eval tasks are effectively infeasible or not reliably evaluable under the evaluation environment (e.g., evidence that is inaccessible or outdated, context-length and tool-output limits, or brittle graders that reject semantically correct answers). Environment-limited failures of this kind cap the attainable scores and make the numbers above conservative, indicating that on the tasks that are in fact solvable RTI has essentially reached the ceiling achievable under the evaluation environment. We attribute this transfer mainly to improved reflection and reasoning: the corrected trajectories teach the model to detect inconsistencies in its own reasoning and to revise its plan accordingly, rather than memorize task-specific solutions, so the acquired multi-step reflective behavior carries over to tasks it has never seen.

\end{document}